\PassOptionsToPackage{table}{xcolor}
\documentclass{article} % For LaTeX2e
\ifdefined\XeTeXversion\usepackage[OT1]{fontenc}\fi
\usepackage{iclr2027_conference,times}
\usepackage{textcomp}
\ifdefined\DeclareUnicodeCharacter\DeclareUnicodeCharacter{266B}{\textmusicalnote}\fi

\usepackage{amsmath,amsfonts,bm}

\def\eqref#1{equation~\ref{#1}}
\def\1{\bm{1}}

\DeclareMathAlphabet{\mathsfit}{\encodingdefault}{\sfdefault}{m}{sl}
\SetMathAlphabet{\mathsfit}{bold}{\encodingdefault}{\sfdefault}{bx}{n}

\usepackage{hyperref}
\usepackage{url}
\usepackage{multirow}
\usepackage{graphicx}
\usepackage{arydshln}
\usepackage{booktabs}
\usepackage{wrapfig}
\usepackage{needspace}

\usepackage{amsmath}
\usepackage{subcaption}
\usepackage{diagbox}
\usepackage{float}
\usepackage{enumitem}
\usepackage{amsmath,amssymb}
\usepackage[ruled,vlined]{algorithm2e}
\usepackage[table]{xcolor}
\usepackage{fontawesome5}  % Overleaf 已经支持
\usepackage{svg}

\newcommand{\coco}{CoCal}

\iclrfinalcopy
\title{Shared Experience, Separate Learning: Companion Confidence Calibration for LLMs}
\author{
Shiyu Ni$^{1,2,3}$ {\quad} Keping Bi$^{1,2,3}$ \quad Jiafeng Guo$^{1,2,3}$ \quad Yilong Xu$^{1,2,3}$ \\
\;\textbf{Jingtong Wu} \quad \textbf{Zengxin Han} \quad \textbf{Xueqi Cheng}$^{1,2,3}$ \\
$^{1}$ State Key Laboratory of AI Safety \\
$^{2}$ Institute of Computing Technology, Chinese Academy of Sciences \\
$^{3}$ University of Chinese Academy of Sciences \\
{\{nishiyu23z, bikeping, guojiafeng, xuyilong23s, cxq\}@ict.ac.cn} \\
wangzhenlingwu@163.com, zengxin.hanzx@gmail.com \\
}

\begin{document}

\maketitle

\begingroup
\renewcommand{\thefootnote}{}
\footnotetext{
\href{https://github.com/ShiyuNee/Companion-Confidence-Calibration}
{\faGithub~Code}
}
\addtocounter{footnote}{-1}
\endgroup

\vspace{-0.7cm}
\begin{abstract}
Reliable self-assessment is essential for large language models (LLMs), yet they often remain highly confident when their answers are wrong. We study \emph{concurrent confidence calibration}, where confidence is learned alongside capability improvement rather than calibrated only after training. Reinforcement learning from verifiable rewards (RLVR) provides a natural setting for this paradigm, as it continuously produces responses paired with verifiable correctness feedback. Existing concurrent methods, however, learn both capability and confidence through reinforcement learning within shared policy parameters, potentially coupling two fundamentally different learning problems. We instead propose \emph{shared experience but separate learning}: capability and confidence learn from the same trajectories, but through separate optimization mechanisms and parameters. Based on this principle, we introduce \textbf{\coco{} (Companion Confidence Calibration)}, which trains a lightweight companion from rollout hidden states and verifier-derived correctness supervision while leaving task optimization unchanged. Experiments on Qwen3-8B and Qwen3-14B show that \coco{} improves confidence estimation without sacrificing task performance, outperforming both RL-based concurrent methods and matched post-hoc calibration. The learned companion further generalizes across domains and policy shifts, while the benefits of \coco{} persist at both scales. \looseness=-1
\end{abstract}

\vspace{-0.7cm}
\section{Introduction}
\vspace{-0.3cm}

Large language models (LLMs) have become capable of solving complex tasks such as mathematical reasoning and software engineering~\citep{guo2025deepseek,yang2025qwen3,jimenez2024swe,openai2026gpt6astra,anthropic2026fable5}. Yet they still struggle to recognize the limits of their own capabilities, often expressing high confidence in incorrect answers~\citep{xiong2023can,ni2024llms,damani2026beyond,ma2026decoupling,liu2026agentabstain,luo2026agentic}. As LLMs take on more autonomous roles, reliable self-assessment becomes increasingly important for deciding when to answer, abstain, retrieve additional information, or allocate more inference compute. \looseness=-1

\begin{wrapfigure}[15]{r}{0.45\textwidth}
\vspace{-0.5cm}
\centering
\captionsetup{font=small,skip=4pt}
\includegraphics[width=\linewidth]{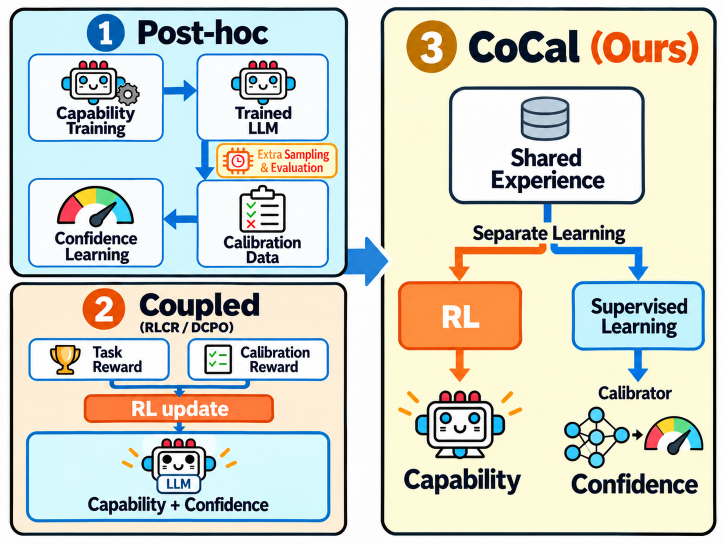}
\caption{Comparison of post-hoc calibration, coupled concurrent training, and \coco{}. \coco{} learns from shared RLVR experience while separating capability and confidence learning.}
\label{fig:paradigm_shift}
\end{wrapfigure}

Human self-assessment co-evolves with capability learning: as people acquire new knowledge and skills, experience continually refines their understanding of what they do and do not know~\citep{flavell1979metacognition,butler1995feedback}. LLM capability training provides a similar opportunity, exposing the model to a broad and evolving range of successes and failures across learning stages. In particular, early errors provide valuable evidence of capability limits but become increasingly scarce as performance improves. Moreover, capability training typically produces large volumes of responses and feedback, which can offer rich and diverse supervision for confidence learning while avoiding the additional sampling and evaluation required by post-hoc calibration~\citep{yang2023alignment,ni2025towards,zhang2024r}. We call this paradigm-learning confidence alongside capability improvement-\emph{concurrent confidence calibration}. \looseness=-1

Extending this paradigm across the full model-training lifecycle would be prohibitively expensive. We therefore focus on RLVR (Reinforcement Learning from Verifiable Rewards)~\citep{guo2025deepseek}, a key post-training stage where responses are naturally paired with verifiable correctness feedback. Recent methods already perform concurrent calibration during RLVR, but formulate confidence learning as part of policy optimization. RLCR~\citep{damani2026beyond} adds confidence-specific reasoning and optimizes the full sequence with combined task and calibration rewards, while DCPO~\citep{ma2026decoupling} reduces interference between the two through token-level objective separation. Despite these differences, both learn confidence through RL within the shared policy parameters.

We propose a different division of labor: \emph{shared experience but separate learning} (Figure~\ref{fig:paradigm_shift}). Task learning needs to improve a policy over possible reasoning trajectories, for which RL is appropriate. Once a response has been generated and verified, however, its correctness provides a direct supervised signal for estimating confidence. Learning this scalar through another policy-gradient objective is not necessary for the formulation, and updating shared parameters gives the two objectives a route to interfere. We therefore reuse the policy's RLVR experience while separating how, and where, confidence is learned.

% We argue for a different principle: \emph{shared experience but separate learning}, as shown in Figure~\ref{fig:paradigm_shift}. Capability and confidence can learn from the same RLVR experience without sharing the same optimization mechanism or parameters. Task learning benefits from exploration over candidate reasoning trajectories, making RL a natural choice. Confidence calibration is fundamentally different: once a response and its correctness are observed, confidence becomes a regression problem that can be learned directly from supervision. Moreover, sharing policy parameters between task and confidence learning may introduce interference between the two objectives.

Based on this division of labor, we propose \textbf{\coco{} (Companion Confidence Calibration)}. Inspired by prior findings that LLM internal representations contain signals predictive of response correctness~\citep{su2024unsupervised,ni2025towards}, \coco{} trains a lightweight companion on rollout hidden states and verifier-derived correctness targets. The companion learns from the same experience as the policy, but with separate parameters and direct supervision, leaving task optimization unchanged.

Experiments on Qwen3-8B and Qwen3-14B yield three main findings.
1) Compared to coupled RL-based methods, \coco{} provides a stronger confidence estimation while preserving task performance; in five mathematical datasets, RLCR and DCPO trail \coco{} in accuracy by 9.9 and 5.0 percentage points on average.
2) Compared with matched post-hoc calibration, \coco{} achieves better confidence estimation while avoiding additional response sampling.
3) \coco{} further widens its advantage over both RL-based concurrent methods and post-hoc calibration under math-to-factual QA transfer. The learned companion also transfers across policy shifts, and \coco{} remains effective on both Qwen3-8B and Qwen3-14B.

\section{Related Work}
\label{sec:related-work}

\noindent\textbf{Unsupervised confidence estimation.}
Some unsupervised methods study how to estimate LLM confidence and can be broadly grouped into three categories.
Probability-based methods derive confidence from token-level generation probabilities~\citep{kadavath2022language,jiang2021can}.
Verbalized methods directly ask models to report their confidence~\citep{lin2022teaching,yin2023large,tian2023just,xiong2023can,ni2024llms}.
Consistency-based methods estimate confidence from agreement across multiple generations, including answer consistency and semantic uncertainty~\citep{manakul2023selfcheckgpt,zhang2023sac3,kuhn2023semantic}.

\noindent\textbf{Post-hoc confidence calibration.}
Many supervised methods calibrate confidence after capability training using response correctness as supervision, either by fine-tuning the model to express calibrated confidence~\citep{yang2023alignment,zhang2024r} or by training separate predictors over internal representations~\citep{azaria2023internal,wang2024hidden,su2024unsupervised,ni2025towards}.
Beyond correctness labels, prior work has incorporated additional signals such as self-consistency~\citep{ni2025annotation}, knowledge popularity~\citep{ni2025knowledge}, and signals from other or base models~\citep{luo2025pretrained,tan2026basecal}.
Reinforcement learning has also been used for post-hoc calibration, including SaySelf~\citep{xu2024sayself} and Rewarding Doubt~\citep{bani2026rewarding}.
Despite these differences, these methods calibrate confidence after the target model has acquired its task capability.

\noindent\textbf{Concurrent confidence calibration.}
More recent work studies confidence learning alongside capability improvement during RLVR.
C$^2$GSPG~\citep{liu2025c} derives confidence from normalized sequence probabilities and jointly optimizes calibration and reasoning within the policy objective.
RLCR~\citep{damani2026beyond} instead learns verbalized confidence through additional confidence-specific reasoning and a joint task--calibration reward, while DCPO~\citep{ma2026decoupling} removes this extra reasoning and separates task and calibration updates across reasoning and confidence tokens.
Despite these differences, capability and confidence remain coupled in both optimization and parameters: both are learned through policy optimization within the same model.
CoCal follows the principle of shared experience but decoupled learning: it reuses RLVR rollouts to directly supervise a separate confidence predictor, while the policy continues to optimize task performance.

\section{Preliminary}
\label{sec:background}

Group Relative Policy Optimization (GRPO)~\citep{shao2024deepseekmath} is a representative algorithm for reinforcement learning with verifiable rewards (RLVR). Given a question $q$, the policy samples a group of $G$ responses $\{o_i\}_{i=1}^{G}$, each receiving a reward $r_i$. GRPO estimates the advantage of each response from its reward relative to others in the same group:
\begin{equation}
A_i = \frac{r_i-\bar{r}}{\sigma_r},
\qquad
\bar{r} = \frac{1}{G}\sum_{j=1}^{G}r_j,
\label{eq:group_advantage}
\end{equation}
where $\sigma_r$ denotes the within-group reward standard deviation. The resulting advantage $A_i$ is applied to the tokens of response $o_i$ through a clipped policy-gradient objective. Following prior RLVR work~\citep{yu2026dapo,damani2026beyond,ma2026decoupling}, we use a rule-based verifier to assess final-answer correctness and use this signal as the task reward. As the policy evolves, its responses and correctness outcomes record successes and failures across different capability stages, providing a natural supervision stream for confidence calibration.

\section{\coco}
\label{sec:method}
\vspace{-0.3cm}

\begin{figure}
    \centering
    \includegraphics[width=\linewidth]{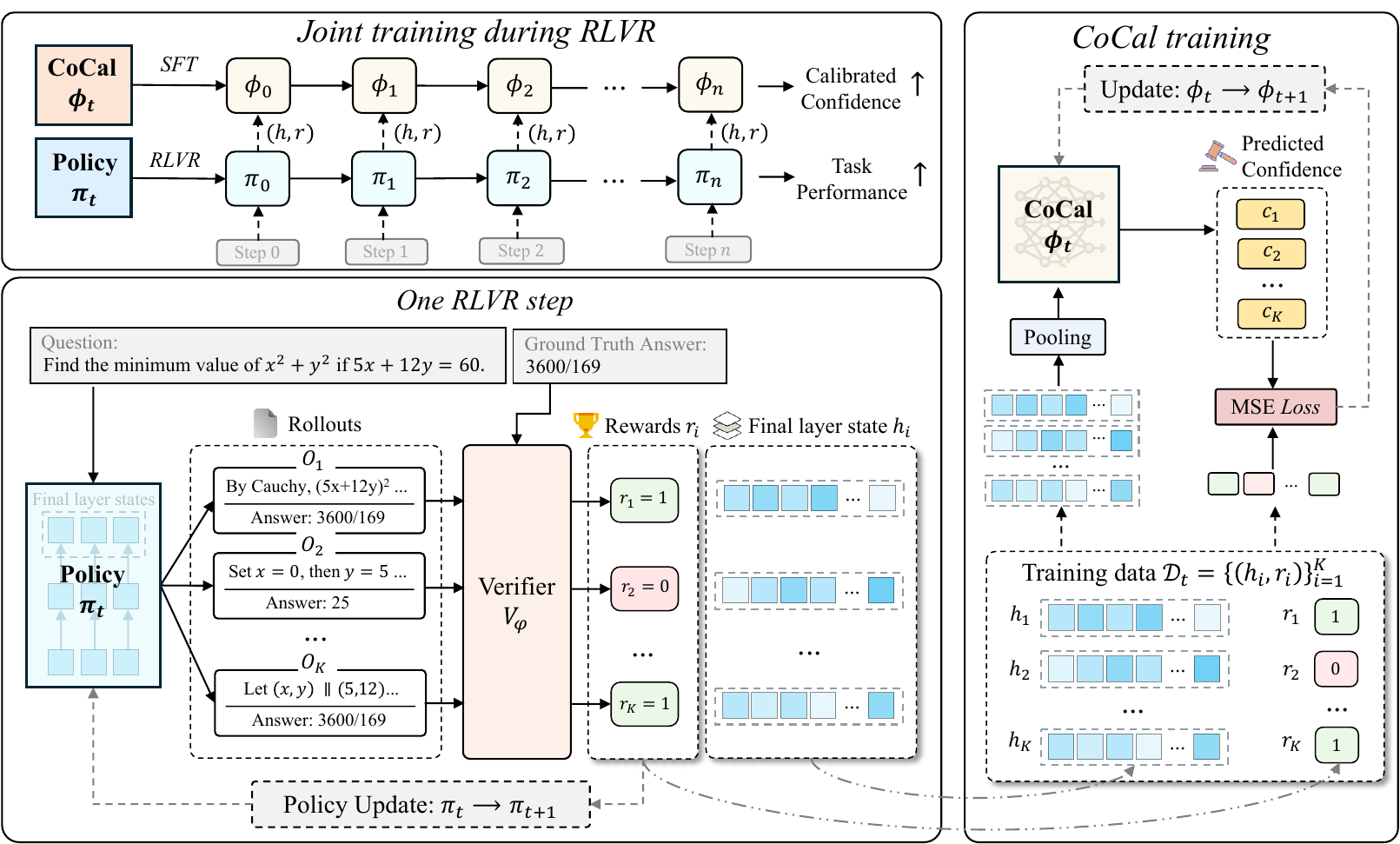}
    \caption{Overview of \coco{}. Shared RLVR rollouts support task learning through policy optimization and confidence learning through direct supervision of a separate companion.}
    \label{fig:cocal_overview}
    \vspace{-0.5cm}
\end{figure}

\coco{} implements \emph{shared experience but decoupled learning} for concurrent confidence calibration. The policy learns task capability through RLVR, while a separate companion learns confidence through direct supervision on the same rollouts. Policy updates and companion updates use separate parameters (Figure~\ref{fig:cocal_overview}).

\subsection{Learning from RLVR Experience}
\label{sec:rollouts}

During RLVR, the policy naturally generates rollouts together with verifier-provided correctness feedback. These signals not only support task learning, but can also provide supervision for confidence learning. \coco{} therefore reuses the rollouts and correctness feedback produced during policy training, avoiding the additional response sampling required by a separate post-hoc calibration stage.

At each training step $t$, the current policy $\pi_t$ receives $B$ questions
$x_1,\ldots,x_B$ and generates $K$ responses for each:
\begin{equation}
    r_{i,j}
    \sim
    \pi_t(\cdot \mid x_i),
    \qquad
    y_{i,j}
    =
    V(x_i,r_{i,j})
    \in\{0,1\},
    \quad j=1,\ldots,K,
\end{equation}
where $V$ is the verifier and $y_{i,j}$ denotes response correctness.

For task learning, the policy is updated by GRPO using response correctness as the reward.
For confidence learning, we use the hidden representation $z_{i,j}$ extracted during response generation as the calibrator input and construct a confidence target $y^{\mathrm{conf}}_{i,j}$ from the verifier feedback.
To prevent confidence learning from interfering with task optimization while keeping the additional module lightweight, we introduce a separate four-layer MLP as the companion calibrator:
\begin{equation}
    c_{i,j}
    =
    g_{\phi_t}(z_{i,j})
    =
    \sigma\!\left(
        \operatorname{MLP}_{\phi_t}(z_{i,j})
    \right)
    \in[0,1].
    \label{eq:companion}
\end{equation}
The calibrator is trained with mean squared error:
\begin{equation}
    \mathcal{L}_{\mathrm{conf}}^{(t)}
    =
    \frac{1}{BK}
    \sum_{i=1}^{B}
    \sum_{j=1}^{K}
    \left(
        c_{i,j}-y^{\mathrm{conf}}_{i,j}
    \right)^2.
    \label{eq:confidence_loss}
\end{equation}

During each chronological pass over the cached data, the companion consumes supervision in the order of the policy steps that produced it:
\begin{equation}
    g_{\phi_0}
    \xrightarrow{\mathcal{S}_0}
    g_{\phi_1}
    \xrightarrow{\mathcal{S}_1}
    \cdots
    \xrightarrow{\mathcal{S}_{T-1}}
    g_{\phi_T},
\end{equation}
where $\mathcal{S}_t=
    \left\{
        \bigl(z_{i,j},y^{\mathrm{conf}}_{i,j}\bigr)
    \right\}_{i,j}$
denotes the confidence-supervision data produced at step $t$.

\paragraph{Practical implementation.}
Because companion optimization is fully decoupled from policy training, the two updates need not be physically interleaved. For implementation convenience, we cache rollout states and verifier labels and train the companion offline by replaying them in chronological order. This provides a practical approximation to online companion learning while preserving supervision from intermediate policy stages, and enables flexible training and evaluation without modifying the RLVR pipeline.

\subsection{Response Representation and Supervision}
\label{sec:companion}

\paragraph{Response representation.}
For a response $r=(w_1,\ldots,w_L)$, let $h_m^{(\ell)}$ denote the hidden
state of its $m$-th token at Transformer layer $\ell$.
Prior work shows that deeper Transformer layers encode richer semantic and
correctness-related information, with final-layer states providing strong
signals for response reliability~\citep{chuang2024dola,ju2024large,su2024unsupervised}.
We therefore extract representation from the final layer.
To capture both global response-level information and fine-grained signals distributed across the response, we concatenate the final-token hidden state with the mean-pooled hidden state.
\begin{equation}
    \bar{h}^{(\ell)}
    =
    \frac{1}{L}
    \sum_{m=1}^{L} h_m^{(\ell)},
    \qquad
    z^{(\ell)}
    =
    h_L^{(\ell)}
    \Vert
    \bar{h}^{(\ell)}.
    \label{eq:response_features}
\end{equation}
Our analysis in~\S\ref{sec:layer_ablation} shows that \coco{} is robust to
token selection and that middle-to-late layers consistently provide strong confidence signals.

\paragraph{Supervision construction.}
For constructing the confidence target $y^{\mathrm{conf}}_{i,j}$ of a given response $r_{i,j}$, we consider two strategies:
\begin{itemize}[leftmargin=0pt,labelsep=0.5em,itemindent=1.2em]

    \item \textbf{Binary supervision (\coco{} (binary)).} 
    We directly supervise whether the response is correct by using the verifier-provided correctness label as the confidence target: $y^{\mathrm{conf}}_{i,j}=y_{i,j}.$
    This provides direct supervision for distinguishing correct from incorrect responses.

    \item \textbf{Fine-grained supervision (\coco{} (group)).}
    To teach the calibrator to distinguish consistently correct responses from occasional successes, we additionally incorporate the success rate of question $x_i$. Let $\bar{y}_i=\frac{1}{K}\sum_{j=1}^{K}y_{i,j}$ denote the group accuracy for question $x_i$. We set $y^{\mathrm{conf}}_{i,j}=\bar{y}_i$ for a correct response ($y_{i,j}=1$), and $y^{\mathrm{conf}}_{i,j}=0$ otherwise. This provides finer-grained supervision among correct responses according to how reliably the policy solves the question.

\end{itemize}

\subsection{Inference and Discussion}
\label{sec:inference}

At inference time, the calibrator predicts confidence from the internal states produced as the policy generates response $r$:
\begin{equation*}
    c(r)=g_{\phi}(z(r)).
\end{equation*}

This design offers several practical advantages. First, confidence learning is separated from policy optimization, avoiding interference with task learning. Second, \coco{} reuses the large and diverse experience naturally produced throughout capability training, which may improve confidence estimation and generalization while eliminating the additional response sampling required by post-hoc calibration. Third, the companion is lightweight, introducing negligible training and inference overhead relative to the LLM (Table~\ref{tab:cost}). The design also entails two trade-offs. \coco{} requires access to internal hidden states and is therefore limited to white-box settings. Moreover, confidence is represented by an external companion rather than internalized into the policy, so \coco{} does not directly improve the model's own verbalized confidence.

\section{Experimental Setup}
\label{sec:experimental_setup}
In this section, we describe the experimental setup, including datasets, models, baselines, evaluation metrics, and training protocol. More details are provided in~\S~\ref{app:implementation_details}.

\paragraph{Datasets.}
We organize the datasets into three groups: training data, mathematical evaluation, and out-of-distribution (OOD) evaluation.

\emph{Training.}
As in~\citep{ma2026decoupling}, we use DeepScaleR~\citep{deepscaler2025}, which contains approximately 40K mathematical problems with reference answers.

\emph{Evaluation on mathematics.}
We evaluate on five mathematical benchmarks spanning a range of difficulty levels.
MATH500~\citep{lightman2024let} contains 500 problems across seven mathematical subjects and five difficulty levels, providing broad coverage from relatively basic to challenging reasoning.
AMC 2023~\citep{amc23} and AMC 2024~\citep{amc24} contain 40 and 45 problems, respectively, drawn from high-school mathematics competitions with a more concentrated competition-level difficulty.
AIME 2024~\citep{aime24} and AIME 2025~\citep{aime25} each contain 30 more challenging invitational problems that typically require deeper reasoning.

\emph{Evaluation on factual QA.}
We evaluate cross-domain transfer on the OOD split of HonestyBench~\citep{ni2025annotation}, a large-scale factual QA benchmark with over 560K training examples and about 70K evaluation examples. We use five datasets from HonestyBench-Eval (hereafter HonestyBench): the single-hop datasets SQuAD~\citep{rajpurkar2016squad} and WebQuestions~\citep{berant-etal-2013-semantic}, the multi-hop datasets ComplexWebQuestions (CWQ)~\citep{talmor2018web} and MuSiQue~\citep{trivedi2022musique}, and the factual-triple-based PopQA~\citep{mallen2022not}.

\paragraph{LLMs and baselines.}
We use Qwen3-8B and Qwen3-14B~\citep{yang2025qwen3} as backbone models in non-thinking mode. We compare \coco{} with seven baselines spanning training-free estimation, post-hoc calibration, and concurrent calibration. More details are provided in Appendix~\ref{app:baseline_details}.
\begin{itemize}[leftmargin=*,itemsep=2pt]

\item \textit{Training-free estimation.}
\textbf{Verbal}~\citep{ni2024llms} directly prompts the model to report a confidence score together with its answer.
\textbf{SelfConsis}~\citep{manakul2023selfcheckgpt} estimates confidence from agreement across multiple sampled responses.
\textbf{TokenProb} derives confidence from the token-level generation probability of the response~\citep{kuhn2023semantic}.
\textbf{P(True)}~\citep{kadavath2022language} asks the model to judge whether its own response is correct and uses the probability assigned to ``True'' as confidence.

\item \textit{Post-hoc calibration.}
\textbf{PostCal}~\citep{ni2025towards} learns confidence after capability training.
Using the final trained policy, it samples the same questions and the same number of responses per question as \coco{} to construct the training data.
It uses the same architecture and targets, with its training configuration detailed in Appendix~\ref{app:baseline_details}. \looseness=-1

\item \textit{Concurrent calibration.}
\textbf{RLCR}~\citep{damani2026beyond} adds confidence-specific reasoning after the answer and jointly optimizes task correctness and confidence calibration with a combined reward.
\textbf{DCPO}~\citep{ma2026decoupling} removes this additional reasoning and separates task and calibration optimization at the token level, applying different advantages to reasoning-and-answer tokens and confidence tokens.
Both methods learn confidence alongside capability improvement through policy optimization within the same model.

\end{itemize}

\paragraph{Metrics.}
We use accuracy to evaluate task performance, expected calibration error (ECE)~\citep{guo2017calibration} to measure confidence calibration, and the area under the receiver operating characteristic curve (AUROC)~\citep{fawcett2006introduction} to assess how well confidence distinguishes correct from incorrect responses. Higher accuracy and AUROC indicate better performance, while lower ECE indicates better calibration. Formal definitions are provided in Appendix~\ref{app:metrics}.

\paragraph{Evaluation Protocol.}
To reduce generation randomness, we sample ten responses per question on datasets with few questions, including AIME24/25 and AMC23/24. For all other datasets, we evaluate a single response per question.
Accuracy is computed over all responses, while confidence metrics are computed
only for responses with valid confidence scores.
For \coco, we independently train and evaluate the calibrator with five random seeds and report the average results. \looseness=-1

\section{Results and Analysis}
In this section, we aim to answer:
(1) within concurrent confidence calibration, does decoupled companion learning outperform optimizing confidence through RL within the shared policy;
(2) what advantages does concurrent calibration offer over post-hoc calibration; and
(3) how well does the learned companion generalize across domains and policy shifts?
We further analyze the effects of model scale, token aggregation, representation layer selection, and pre-generation confidence estimation, with full results provided in Appendix~\S\ref{app:model_scale}--\ref{app:pre_generation}.

% ============================================================
\subsection{\coco{} Outperforms Existing Concurrent Methods}
\label{sec:concurrent_comparison}

We first compare \coco{} with training-free and RL-based concurrent confidence estimation methods. The comparison with PostCal is discussed separately in~\S\ref{sec:postcal}. Results are shown in Table~\ref{tab:main_iter119}. We find:

\begin{table}[h]
\centering
\captionsetup{font=small,skip=4pt}
\caption{Accuracy and confidence quality on five mathematical datasets using Qwen3-8B after 120 training steps. Base + Verbal uses the pre-RL model; other training-free baselines use the GRPO-trained model. Bold and underlined values indicate the best and second-best results, respectively.}
\label{tab:main_iter119}
\setlength{\tabcolsep}{2.6pt}
\renewcommand{\arraystretch}{1.08}
\resizebox{\textwidth}{!}{%
\begin{tabular}{l*{12}{c}}
\toprule
\multirow{2}{*}{Method}
& \multicolumn{4}{c}{MATH500}
& \multicolumn{4}{c}{AIME24}
& \multicolumn{4}{c}{AIME25} \\
\cmidrule(lr){2-5}\cmidrule(lr){6-9}\cmidrule(lr){10-13}
& Acc.$\uparrow$ & MeanConf & ECE$\downarrow$ & AUROC$\uparrow$
& Acc.$\uparrow$ & MeanConf & ECE$\downarrow$ & AUROC$\uparrow$
& Acc.$\uparrow$ & MeanConf & ECE$\downarrow$ & AUROC$\uparrow$ \\
\midrule
Base + Verbal & 79.8 & 0.995 & 0.196 & 0.566 & 25.0 & 0.972 & 0.718 & 0.587 & 18.7 & 0.977 & 0.790 & 0.730 \\
\midrule
Verbal & 85.2 & 0.997 & 0.059 & 0.548 & \underline{43.0} & 0.956 & 0.498 & 0.877 & \underline{33.0} & 0.963 & 0.646 & 0.865 \\
SelfConsis & \underline{86.0} & 0.868 & \textbf{0.026} & \textbf{0.945} & \textbf{47.3} & 0.576 & 0.118 & \textbf{0.937} & \textbf{34.7} & 0.603 & 0.257 & 0.869 \\
TokenProb & \underline{86.0} & 0.819 & 0.087 & 0.834 & \textbf{47.3} & 0.668 & 0.246 & 0.873 & \textbf{34.7} & 0.638 & 0.330 & 0.943 \\
P(True) & \underline{86.0} & 0.988 & 0.128 & 0.861 & \textbf{47.3} & 0.960 & 0.486 & 0.908 & \textbf{34.7} & 0.981 & 0.634 & 0.927 \\
\midrule
RLCR & 85.6 & 0.908 & 0.064 & 0.797 & 32.3 & 0.338 & \textbf{0.087} & 0.906 & 28.3 & 0.333 & \textbf{0.050} & \underline{0.965} \\
DCPO & \textbf{86.2} & 0.956 & 0.072 & 0.687 & 39.3 & 0.532 & 0.157 & \underline{0.926} & 30.7 & 0.492 & 0.191 & 0.956 \\
PostCal (group) & \underline{86.0} & 0.849 & 0.044 & 0.870 & \textbf{47.3} & 0.446 & 0.133 & 0.910 & \textbf{34.7} & 0.424 & 0.158 & 0.948 \\
PostCal (binary) & \underline{86.0} & 0.887 & 0.049 & 0.754 & \textbf{47.3} & 0.598 & 0.134 & 0.894 & \textbf{34.7} & 0.566 & 0.245 & \textbf{0.967} \\
\midrule
\coco{} (group) & \underline{86.0} & 0.826 & 0.049 & 0.911 & \textbf{47.3} & 0.429 & \underline{0.100} & 0.910 & \textbf{34.7} & 0.408 & \underline{0.128} & 0.936 \\
\textbf{\coco{} (binary)} & \underline{86.0} & 0.843 & \underline{0.038} & \underline{0.912} & \textbf{47.3} & 0.505 & 0.102 & 0.910 & \textbf{34.7} & 0.456 & 0.147 & 0.944 \\
\bottomrule
\end{tabular}%
}

\vspace{2pt}

\resizebox{\textwidth}{!}{%
\begin{tabular}{l*{12}{c}}
\toprule
\multirow{2}{*}{Method}
& \multicolumn{4}{c}{AMC23}
& \multicolumn{4}{c}{AMC24}
& \multicolumn{4}{c}{Macro Average} \\
\cmidrule(lr){2-5}\cmidrule(lr){6-9}\cmidrule(lr){10-13}
& Acc.$\uparrow$ & MeanConf & ECE$\downarrow$ & AUROC$\uparrow$
& Acc.$\uparrow$ & MeanConf & ECE$\downarrow$ & AUROC$\uparrow$
& Acc.$\uparrow$ & MeanConf & ECE$\downarrow$ & AUROC$\uparrow$ \\
\midrule
Base + Verbal & 69.8 & 0.991 & 0.297 & 0.633 & 43.3 & 0.980 & 0.539 & 0.638 & 47.3 & 0.983 & 0.508 & 0.631 \\
\midrule
Verbal & 74.0 & 0.984 & 0.209 & 0.633 & \textbf{65.8} & 0.969 & 0.282 & 0.692 & \underline{60.2} & 0.974 & 0.339 & 0.723 \\
SelfConsis & \textbf{85.8} & 0.846 & \textbf{0.064} & \textbf{0.967} & \underline{65.1} & 0.789 & 0.181 & 0.801 & \textbf{63.8} & 0.736 & 0.129 & \textbf{0.904} \\
TokenProb & \textbf{85.8} & 0.744 & 0.131 & 0.848 & \underline{65.1} & 0.716 & \underline{0.066} & 0.718 & \textbf{63.8} & 0.717 & 0.172 & 0.843 \\
P(True) & \textbf{85.8} & 0.991 & 0.134 & 0.770 & \underline{65.1} & 0.982 & 0.331 & 0.830 & \textbf{63.8} & 0.980 & 0.343 & 0.859 \\
\midrule
RLCR & 63.5 & 0.724 & 0.191 & 0.769 & 59.6 & 0.681 & 0.133 & 0.848 & 53.9 & 0.597 & 0.105 & 0.857 \\
DCPO & \underline{79.0} & 0.814 & 0.122 & 0.779 & 58.9 & 0.714 & 0.152 & 0.806 & 58.8 & 0.702 & 0.139 & 0.831 \\
PostCal (group) & \textbf{85.8} & 0.684 & 0.185 & 0.852 & \underline{65.1} & 0.624 & 0.080 & 0.802 & \textbf{63.8} & 0.605 & 0.120 & 0.876 \\
PostCal (binary) & \textbf{85.8} & 0.794 & \underline{0.083} & \underline{0.878} & \underline{65.1} & 0.710 & 0.090 & \textbf{0.867} & \textbf{63.8} & 0.711 & 0.120 & 0.872 \\
\midrule
\coco{} (group) & \textbf{85.8} & 0.681 & 0.176 & 0.853 & \underline{65.1} & 0.619 & \textbf{0.065} & 0.808 & \textbf{63.8} & 0.593 & \underline{0.103} & 0.884 \\
\textbf{\coco{} (binary)} & \textbf{85.8} & 0.740 & 0.118 & 0.866 & \underline{65.1} & 0.676 & 0.077 & \underline{0.851} & \textbf{63.8} & 0.644 & \textbf{0.096} & \underline{0.896} \\
\bottomrule
\end{tabular}%
}
\end{table}

\noindent\textbf{\coco{} achieves strong confidence estimation with negligible additional inference overhead.}
As shown in Table~\ref{tab:main_iter119}, \coco{} (binary) achieves the lowest ECE (0.096) among concurrent methods and an AUROC of 0.896, outperforming RLCR (0.105/0.857) and DCPO (0.139/0.831).
This advantage emerges early: at step 40, \coco{} already reaches an ECE of around 0.10, while RLCR and DCPO remain relatively weak and improve substantially only by step 80; even the first 10 rollout steps provide effective supervision for useful confidence estimation (Appendix Figure~\ref{fig:early_step_cocal}).
Training-free methods are generally overconfident.
SelfConsis is the strongest among them, with comparable AUROC but higher ECE and the cost of ten generations per question.
As shown in Table~\ref{tab:cost}, a single LLM generation takes 0.372 seconds on average, whereas \coco{} adds only 0.199\,ms per response.

\begin{figure}[!ht]
% \vspace{-0.3cm}
  \centering
  \captionsetup{font=small,skip=4pt}
  \includegraphics[width=0.98\textwidth]{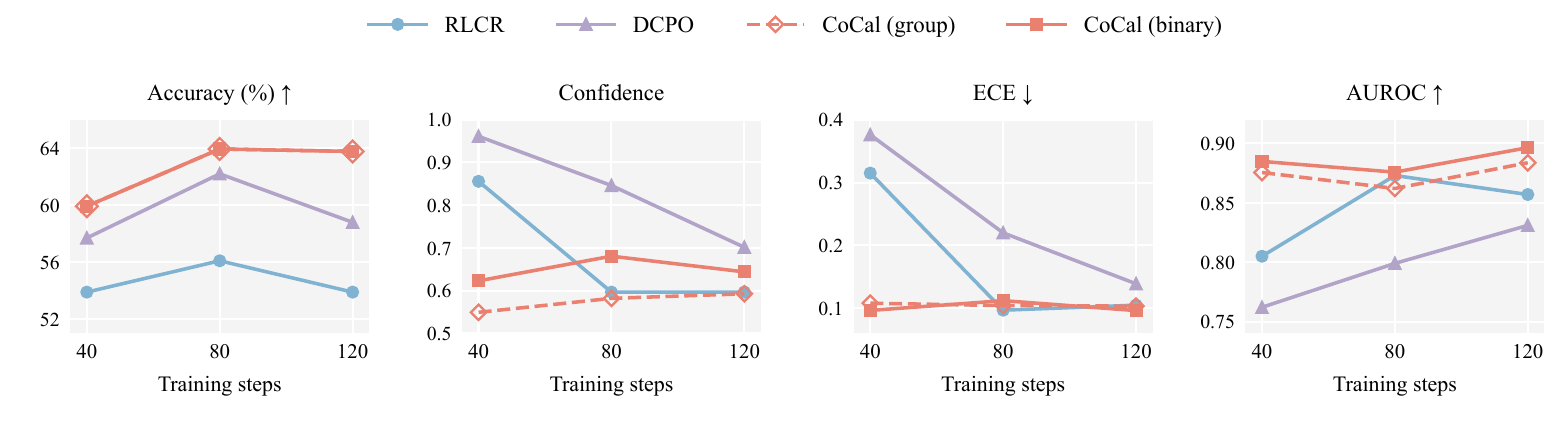}
  \caption{Task accuracy, confidence, calibration, and discrimination across policy-training stages on five mathematical datasets. CoCal results average five seeds; both variants share the same task accuracy.}
  \label{fig:training_dynamics}
  \vspace{-0.5cm}
\end{figure}

\noindent\textbf{Coupled concurrent confidence calibration can interfere with task learning.}
Figure~\ref{fig:training_dynamics} compares RLCR and DCPO with \coco{} across training stages.
RLCR and DCPO consistently achieve lower task accuracy, ending 9.9 and 5.0 percentage points below \coco{}, respectively.
RLCR attains stronger confidence estimation than DCPO but at a larger accuracy cost, suggesting a capability--calibration trade-off under shared policy optimization.
This difference is consistent with their optimization scopes: RLCR applies confidence-related optimization to the full sequence, whereas DCPO restricts it to confidence tokens.
Both methods also show declining accuracy as confidence estimation improves, while \coco{} remains comparatively stable.

\begin{figure}[!h]
  \centering
  \vspace{-0.4cm}
  \captionsetup{font=small,skip=4pt}
  \includegraphics[width=0.98\textwidth]{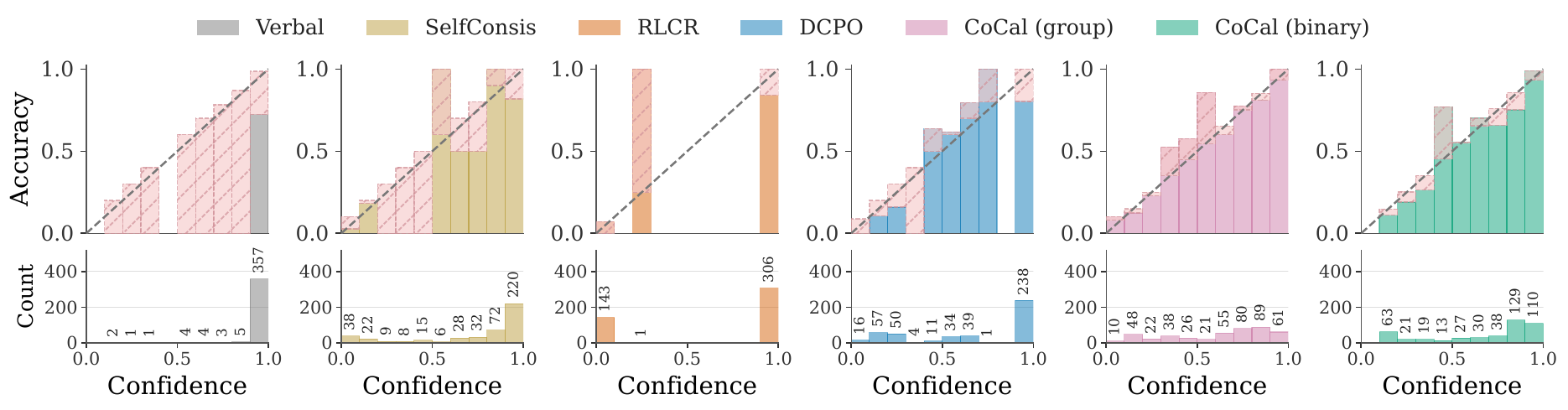}
  \caption{Reliability diagrams and confidence histograms on AMC24 after 120 training steps. Upper panels show bin accuracy and the ideal-calibration diagonal; lower panels show response counts.}
  \label{fig:rq2_reliability}
  \vspace{-0.4cm}
\end{figure}

\noindent\textbf{Direct supervision yields more fine-grained confidence estimates.}
Figure~\ref{fig:rq2_reliability} shows that RLCR produces highly polarized confidence estimates on AMC24, assigning over 99\% of responses to the lowest or highest bin, while DCPO remains concentrated at high confidence.
In contrast, both \coco{} variants spread predictions across a broader range.
Binary supervision directly targets response correctness and performs better overall on mathematics, whereas group supervision incorporates question-level reliability and produces more conservative estimates.
Corresponding results for all five datasets are provided in Appendix Figure~\ref{fig:rq2_reliability_all}.
\looseness=-1

% ============================================================
% \Needspace{17\baselineskip}
\subsection{Benefits of \coco{} over Post-hoc Calibration}
\label{sec:postcal}

We next examine whether confidence is better learned alongside capability improvement.
We compare \coco{} with PostCal, which uses the same calibrator architecture and supervision but learns from newly sampled responses of the final policy.
Table~\ref{tab:main_iter119} reports confidence performance on mathematics, while Table~\ref{tab:cost} compares computational cost. Detailed settings are provided in Appendix~\S\ref{app:comparision_with_postcal}.

\noindent\textbf{Concurrent calibration yields better confidence estimates.}
On the mathematical benchmarks in Table~\ref{tab:main_iter119}, both \coco{} variants achieve lower ECE and higher AUROC than their PostCal counterparts. This advantage is consistent across matched training data budgets: \coco{} obtains lower ECE and higher AUROC in seven of eight settings (Appendix Figure~\ref{fig:rq4_cocal_vs_posthoc}). The performance gap widens further under math-to-factual transfer, as discussed in~\S\ref{sec:OOD}, consistent with the benefit of learning confidence from broader experience across capability-learning stages.\looseness=-1

\begin{wraptable}[11]{r}{0.48\textwidth}
\vspace{-0.4cm}
\centering
\captionsetup{font=small,skip=4pt}
\small
\setlength{\tabcolsep}{1.8pt}
\renewcommand{\arraystretch}{1.08}
\caption{Training and per-question inference costs. LLM inference latency is averaged over MATH500 questions using vLLM on a B200 GPU.}
\label{tab:cost}
\begin{tabular}{lccccc}
\toprule
\multirow{3}{*}{Method}
& \multicolumn{3}{c}{Training}
& \multicolumn{2}{c}{Inference} \\
\cmidrule(lr){2-4}\cmidrule(lr){5-6}
& LLM & Sampling & MLP & LLM & MLP \\
& (h) & (h) & (s) & (s) & (ms) \\
\midrule
\coco{}  & 16.67 & --    & 15.1 & 0.372 & 0.199 \\
PostCal & 16.67 & 14.22 & $\approx15.1$  & 0.372 & 0.199 \\
SelfConsis & 16.67 & -- & -- & 3.225 & -- \\
\bottomrule
\end{tabular}
\end{wraptable}

\noindent\textbf{\coco{} avoids additional response sampling.}
\coco{} reuses rollouts already produced during policy training, whereas PostCal needs to sample new additional responses from the model after capability training. This extra sampling takes 14.22 hours with vLLM~\citep{kwon2023efficient} on a single B200 GPU (Table~\ref{tab:cost}). By contrast, the confidence predictor itself is lightweight, requiring only about 15\,s of training and 0.199\,ms per-response inference, compared with 0.372\,s for a single LLM generation. Detailed cost settings are provided in Appendix~\S\ref{app:cost}.

\begin{table}[!h]
\centering
\vspace{-0.3cm}
\captionsetup{font=small,skip=4pt}
\caption{Math-to-factual transfer on HonestyBench using Qwen3-8B after 120 training steps. Base + Verbal uses the pre-RL model; other training-free baselines use the GRPO-trained model. Bold and underlined values indicate the best and second-best results, respectively.}
\label{tab:honestybench_ood_iter119_8b_all}
\setlength{\tabcolsep}{2.6pt}
\renewcommand{\arraystretch}{1.08}
\resizebox{\textwidth}{!}{%
\begin{tabular}{l*{12}{c}}
\toprule
\multirow{2}{*}{Method}
& \multicolumn{4}{c}{SQuAD}
& \multicolumn{4}{c}{WebQuestions}
& \multicolumn{4}{c}{ComplexWebQuestions} \\
\cmidrule(lr){2-5}\cmidrule(lr){6-9}\cmidrule(lr){10-13}
& Acc.$\uparrow$ & MeanConf & ECE$\downarrow$ & AUROC$\uparrow$
& Acc.$\uparrow$ & MeanConf & ECE$\downarrow$ & AUROC$\uparrow$
& Acc.$\uparrow$ & MeanConf & ECE$\downarrow$ & AUROC$\uparrow$ \\
\midrule
Base + Verbal & 30.7 & 0.923 & 0.620 & 0.647 & 53.9 & 0.983 & 0.442 & 0.608 & 33.2 & 0.947 & 0.625 & 0.632 \\
\midrule
Verbal & 31.1 & 0.907 & 0.603 & 0.656 & 53.7 & 0.978 & 0.441 & 0.623 & 33.0 & 0.918 & 0.596 & 0.677 \\
SelfConsis & \textbf{31.7} & 0.749 & 0.432 & 0.687 & \textbf{57.1} & 0.755 & 0.211 & \textbf{0.712} & \textbf{34.2} & 0.643 & 0.301 & \underline{0.754} \\
TokenProb & \textbf{31.7} & 0.871 & 0.554 & 0.667 & \textbf{57.1} & 0.883 & 0.313 & 0.648 & \textbf{34.2} & 0.881 & 0.539 & \textbf{0.759} \\
P(True) & \textbf{31.7} & 0.845 & 0.554 & 0.666 & \textbf{57.1} & 0.886 & 0.374 & 0.643 & \textbf{34.2} & 0.809 & 0.501 & 0.694 \\
\midrule
RLCR & 30.6 & 0.570 & 0.284 & \underline{0.741} & \underline{56.1} & 0.898 & 0.345 & 0.601 & \underline{33.8} & 0.603 & 0.354 & 0.677 \\
DCPO & \underline{31.5} & 0.751 & 0.444 & 0.712 & 54.8 & 0.939 & 0.392 & 0.619 & 33.0 & 0.767 & 0.455 & 0.672 \\
PostCal (group) & \textbf{31.7} & 0.344 & 0.143 & 0.668 & \textbf{57.1} & 0.448 & 0.237 & 0.629 & \textbf{34.2} & 0.315 & 0.155 & 0.645 \\
PostCal (binary) & \textbf{31.7} & 0.431 & 0.176 & 0.677 & \textbf{57.1} & 0.550 & 0.236 & 0.648 & \textbf{34.2} & 0.365 & 0.164 & 0.651 \\
\midrule
\coco{} (group) & \textbf{31.7} & 0.352 & \textbf{0.037} & 0.738 & \textbf{57.1} & 0.442 & \underline{0.143} & \underline{0.700} & \textbf{34.2} & 0.331 & \textbf{0.032} & 0.707 \\
\textbf{\coco{} (binary)} & \textbf{31.7} & 0.391 & \underline{0.074} & \textbf{0.747} & \textbf{57.1} & 0.544 & \textbf{0.130} & \underline{0.700} & \textbf{34.2} & 0.352 & \underline{0.044} & 0.710 \\
\bottomrule
\end{tabular}%
}

\vspace{2pt}

\resizebox{\textwidth}{!}{%
\begin{tabular}{l*{12}{c}}
\toprule
\multirow{2}{*}{Method}
& \multicolumn{4}{c}{MuSiQue}
& \multicolumn{4}{c}{PopQA}
& \multicolumn{4}{c}{Macro Average} \\
\cmidrule(lr){2-5}\cmidrule(lr){6-9}\cmidrule(lr){10-13}
& Acc.$\uparrow$ & MeanConf & ECE$\downarrow$ & AUROC$\uparrow$
& Acc.$\uparrow$ & MeanConf & ECE$\downarrow$ & AUROC$\uparrow$
& Acc.$\uparrow$ & MeanConf & ECE$\downarrow$ & AUROC$\uparrow$ \\
\midrule
Base + Verbal & \textbf{16.1} & 0.849 & 0.770 & 0.468 & 22.1 & 0.953 & 0.732 & 0.543 & \underline{31.2} & 0.931 & 0.638 & 0.580 \\
\midrule
Verbal & 15.3 & 0.793 & 0.699 & 0.534 & \underline{22.4} & 0.946 & 0.723 & 0.569 & 31.1 & 0.908 & 0.612 & 0.612 \\
SelfConsis & 11.3 & 0.503 & 0.390 & \textbf{0.747} & 21.5 & 0.490 & 0.275 & \textbf{0.794} & 31.1 & 0.628 & 0.322 & \textbf{0.739} \\
TokenProb & 11.3 & 0.846 & 0.733 & 0.722 & 21.5 & 0.870 & 0.655 & \underline{0.788} & 31.1 & 0.870 & 0.559 & 0.717 \\
P(True) & 11.3 & 0.787 & 0.678 & 0.718 & 21.5 & 0.749 & 0.550 & 0.747 & 31.1 & 0.815 & 0.531 & 0.694 \\
\midrule
RLCR & 13.4 & 0.323 & 0.289 & 0.649 & 22.2 & 0.702 & 0.497 & 0.662 & \underline{31.2} & 0.619 & 0.354 & 0.666 \\
DCPO & \underline{15.9} & 0.548 & 0.468 & 0.535 & \textbf{22.8} & 0.884 & 0.658 & 0.582 & \textbf{31.6} & 0.778 & 0.483 & 0.624 \\
PostCal (group) & 11.3 & 0.233 & \textbf{0.127} & 0.662 & 21.5 & 0.400 & 0.218 & 0.654 & 31.1 & 0.348 & 0.176 & 0.651 \\
PostCal (binary) & 11.3 & 0.268 & 0.155 & 0.661 & 21.5 & 0.508 & 0.299 & 0.701 & 31.1 & 0.425 & 0.206 & 0.667 \\
\midrule
\coco{} (group) & 11.3 & 0.264 & \underline{0.151} & \underline{0.729} & 21.5 & 0.375 & \textbf{0.161} & 0.751 & 31.1 & 0.353 & \textbf{0.104} & 0.725 \\
\textbf{\coco{} (binary)} & 11.3 & 0.269 & 0.156 & \underline{0.729} & 21.5 & 0.422 & \underline{0.207} & 0.745 & 31.1 & 0.396 & \underline{0.122} & \underline{0.726} \\
\bottomrule
\end{tabular}%
}
\end{table}

% ============================================================
\vspace{-0.5cm}
\subsection{\coco{} Generalizes across Domains and Policies}
\label{sec:OOD}

We assess the generalization of \coco{} along two complementary dimensions.
Cross-domain transfer tests whether confidence learned from mathematical reasoning transfers to factual QA, while cross-policy transfer tests whether the learned companion can be reused across different models.

\noindent\textbf{\coco{} gains a larger advantage under domain shift.}
Table~\ref{tab:honestybench_ood_iter119_8b_all} shows that \coco{} transfers effectively from mathematical reasoning to factual QA, and its advantage over both RL-based concurrent methods and PostCal becomes substantially larger under domain shift.
With group supervision, \coco{} achieves an ECE of 0.104, compared with 0.354 for RLCR, 0.483 for DCPO, and 0.176 for PostCal.
With binary supervision, \coco{} reaches an AUROC of 0.726, compared with 0.666, 0.624, and 0.667, respectively.
The widening gap relative to PostCal is consistent with the benefit of broader supervision across capability-learning stages, which may improve cross-domain generalization of the learned confidence.
Among training-free methods, SelfConsis retains strong discrimination with an AUROC of 0.739, but its ECE rises to 0.322.
Verbal, TokenProb, and P(True) show similar overconfidence.
Overall, \coco{} remains well calibrated while retaining competitive discrimination under substantial domain shift.

\begin{figure*}[h]
\centering
\vspace{-0.7cm}

\begin{minipage}[t]{0.48\textwidth}
\vspace{0pt}
\centering

\begin{minipage}[c][4.2cm][c]{\linewidth}
    \centering
    \includegraphics[
        width=\linewidth,
        height=4.0cm,
        keepaspectratio
    ]{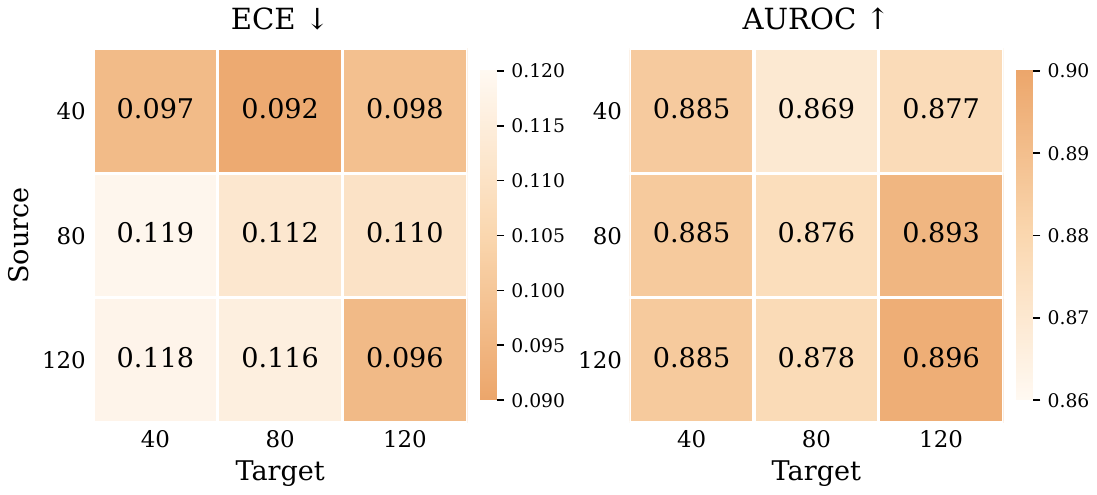}
\end{minipage}

\captionsetup{font=small,skip=4pt}
\caption{Cross-iteration transfer of \coco{}. Source indicates the rollout-data prefix used for companion training, and target the policy used for evaluation.}
\label{fig:rq3_cross_iter}
\end{minipage}
\hfill
\begin{minipage}[t]{0.48\textwidth}
\vspace{0pt}
\centering

\begin{minipage}[c][4.2cm][c]{\linewidth}
    \centering
    \includegraphics[
        width=\linewidth,
        height=4.0cm,
        keepaspectratio
    ]{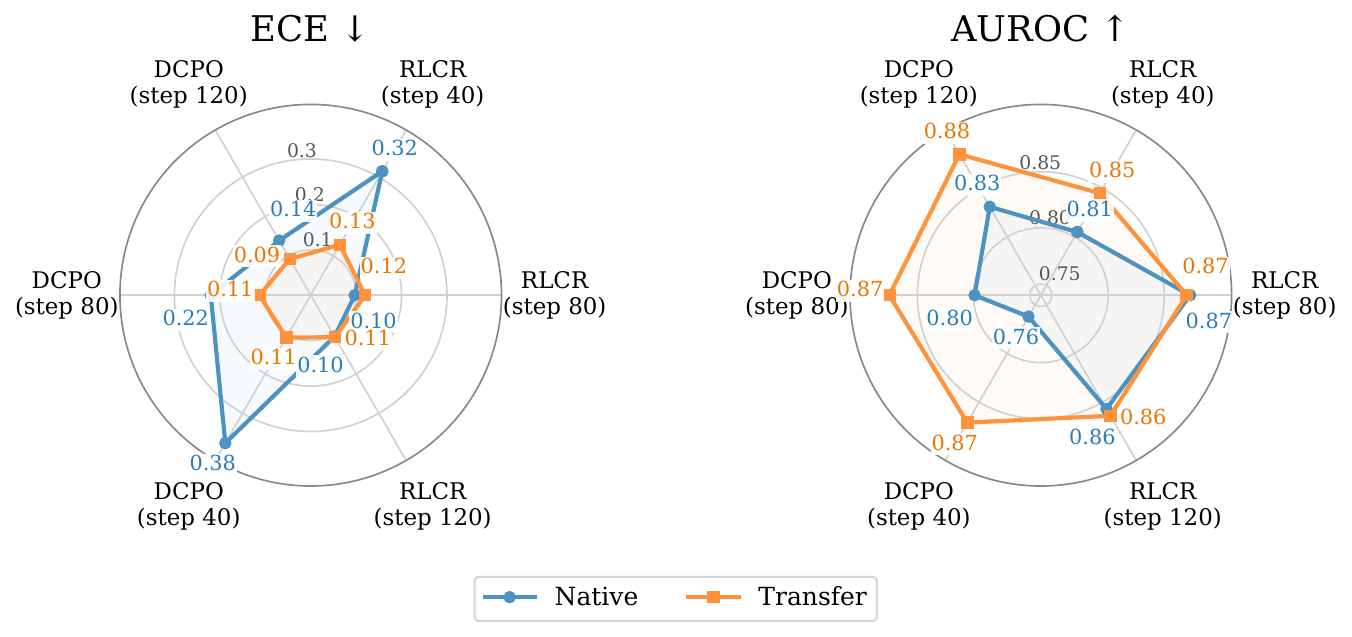}
\end{minipage}

\captionsetup{font=small,skip=4pt}
\caption{Cross-policy transfer of \coco{} across training methods. The calibrator is applied to models under DCPO and RLCR after the same training steps.}
\label{fig:rq3_policy_transfer}
\end{minipage}
\end{figure*}

\noindent\textbf{\coco{} also transfers across policy changes.}
Figure~\ref{fig:rq3_cross_iter} evaluates transfer between policies from different stages of capability learning using \coco{} (binary).
Across the six off-diagonal source--target pairs, AUROC remains between 0.869 and 0.893 and ECE between 0.092 and 0.119, showing that the learned confidence signal remains stable as the policy evolves.
Figure~\ref{fig:rq3_policy_transfer} further evaluates transfer to RLCR- and DCPO-trained policies.
Transfer improves both ECE and AUROC across all three DCPO stages and for RLCR at step 40, but becomes less consistent for later RLCR policies.
A likely reason is that RLCR alters the generation process more substantially by introducing confidence-specific reasoning, creating a larger shift in internal representations.
Overall, \coco{} transfers reliably across training stages and remains useful under independently optimized policies, although larger representation shifts can reduce transfer quality.

% ============================================================
\subsection{Further Analysis}
\label{sec:layer_ablation}

We further analyze \coco{} along four dimensions: model scale, token aggregation, representation layer, and pre-generation confidence estimation. \coco{} remains effective when scaling from Qwen3-8B to Qwen3-14B and is robust to different token aggregation strategies. Middle-to-late Transformer layers provide similarly strong confidence signals, suggesting that there is no need for careful layer selection. Prompt-only states also support informative pre-generation estimates; although less accurate than post-generation confidence, they reduce time to confidence by roughly $10^2$--$10^3\times$ by avoiding response generation. Due to space constraints, full results are provided in Appendix~\S\ref{app:model_scale}-\ref{app:pre_generation}.

\section{Conclusion}
We introduced \coco{}, a companion approach to concurrent confidence calibration built on the principle of \emph{shared experience but separate learning}. Rather than learning confidence through policy optimization, \coco{} trains a lightweight predictor directly from RLVR rollouts and verifier feedback while leaving task learning unchanged. Across Qwen3-8B and Qwen3-14B, this simple separation yields strong calibration and discrimination without sacrificing task accuracy, with useful confidence signals emerging from early training experience and transferring across domains and policy states. \coco{} also generally outperforms post-hoc calibration while eliminating the need for additional response sampling after capability training. These results suggest that confidence need not be entangled with capability optimization: the two can co-evolve from the same experience through separate learning processes. \looseness=-1

\section*{Ethics Statement}
This work uses publicly available models and datasets for mathematical reasoning and factual question answering. Our research focuses on improving confidence estimation in large language models to support more reliable assessment of their responses.

\section*{Reproducibility Statement}
We provide the experimental details needed to facilitate reproduction of our main results. Section~5 describes the experimental setup, baselines, and evaluation protocol, while Appendix~\ref{app:implementation_details} provides implementation details including model architecture, training hyperparameters, data splits, and random seeds. Additional experimental settings and results are provided in Appendices~C--F.

\section*{AI Use Statement}
Generative AI tools were used for language editing, figure presentation, manuscript organization, and refining the presentation of experimental analyses. The research motivation, methodology, experimental design, implementation, and experimental results were produced by the authors. All AI-assisted text and presentation suggestions were reviewed and verified by the authors, who take full responsibility for the final content of the paper.

% \section*{ACKNOWLEDGEMENTS}

% In practice, $y^*(x)$ is usually obtained through greedy search to ensure reproducibility.

% \paragraph{Remark (relation to correctness probability).}
% For completeness, recall the capability-aligned correctness probability
% \(\;q(x)=\mathbb{E}_{y\sim p_\theta^\pi}[\, z(x,y)\,]\;\) where \(z(x,y)=\mathbb{I}[y\in\mathcal{A}(x)]\).
% While \(q(x)\) quantifies the probability that a sampled answer is {\em acceptable/correct}, \(S(x)\) (or \(S_{\mathrm{sem}}(x)\)) quantifies the model's {\em internal confidence} in its most-probable answer (i.e., distributional concentration around \(y^\star\)).  These are distinct: a model may have high self-consistency but low correctness, or vice versa.

\bibliography{iclr2025_conference}
\bibliographystyle{iclr2027_conference}

\appendix
\raggedbottom
\section{Implementation Details}
\label{app:implementation_details}

\subsection{Policy and Companion Training}
Policies are trained with \texttt{slime}~\citep{slime_github} for 120 steps and evaluated every 40 steps~\citep{ma2026decoupling}. Each step samples eight responses for each of 256 questions.

\paragraph{Feature extraction.}
We extract response hidden states from the final Transformer layer before the final RMSNorm, excluding prompt tokens. For calibrator training and validation, we discard responses that reach the maximum generation length, as these truncated outputs are incomplete and may provide unreliable response representations and correctness supervision. For responses that terminate normally, the generated end-of-sequence token is included in the extracted representation.

\paragraph{Calibrator training.}
For practical training, we cache the rollout hidden states and correctness labels produced at each RLVR step and replay them offline in chronological order to emulate step-wise companion learning. For the calibrator associated with policy step $t$, we use all data produced up to that step, ordered by their policy-training steps. 
% To provide sufficient optimization and avoid underfitting, we repeat this chronological traversal for multiple epochs without shuffling.

The MLP uses the widths and optimization settings in Table~\ref{tab:hyperparameters}, with LayerNorm, GELU~\citep{hendrycks2016gaussian}, dropout rates of 0.1/0.1/0.05 after the first three hidden layers, and a sigmoid output. We split questions 90:10 into training and validation sets, keeping all responses to the same question in the same partition. Group targets retain the original eight-response group accuracy before truncation filtering. Checkpoints are selected by validation MSE against binary correctness.

\subsection{Evaluation and Metric Definitions}
\label{app:metrics}
The sampling parameters in Table~\ref{tab:hyperparameters} apply to stochastic generation. Test responses that reach the generation-length limit are retained, although they account for only a small fraction of the evaluation set. For methods that generate verbalized confidence, including Verbal, RLCR, and DCPO, some responses may fail to produce a valid confidence score. We therefore compute accuracy over all evaluated responses, while confidence metrics are computed only over responses with a valid confidence score in $[0,1]$.
For a confidence-valid set of $N$ responses, let $c_i$ and $y_i$ denote predicted confidence and binary correctness. We divide $[0,1]$ into ten equal-width bins $\mathcal{B}_b$: $[0,0.1]$, $(0.1,0.2]$, $\ldots$, $(0.9,1]$. For each bin,
\begin{equation}
\operatorname{acc}(\mathcal{B}_b)=\frac{1}{|\mathcal{B}_b|}\sum_{i\in\mathcal{B}_b}y_i,
\qquad
\operatorname{conf}(\mathcal{B}_b)=\frac{1}{|\mathcal{B}_b|}\sum_{i\in\mathcal{B}_b}c_i.
\end{equation}
The expected calibration error is:
\begin{equation}
\mathrm{ECE}=\sum_{b:\,|\mathcal{B}_b|>0}\frac{|\mathcal{B}_b|}{N}
\left|\operatorname{acc}(\mathcal{B}_b)-\operatorname{conf}(\mathcal{B}_b)\right|.
\end{equation}
AUROC measures the ranking of correct versus incorrect responses. We first compute metrics separately for each dataset and then average them with equal weight. Results for \coco{} average five independently trained companions with randomly sampled seeds 42, 137, 500, 700, and 1024. 

\begin{table}[!htbp]
\centering
\caption{Hyperparameters for policy training, calibrator training, and evaluation. Response lengths are measured in tokens.}
\label{tab:hyperparameters}
\setlength{\tabcolsep}{8pt}
\renewcommand{\arraystretch}{1.05}
\begin{tabular}{lc}
\toprule
Setting & Value \\
\midrule
\multicolumn{2}{l}{\textit{Policy training}} \\
Training steps & 120 \\
Questions per step & 256 \\
Responses per question & 8 \\
Maximum response length (GRPO / DCPO) & 3,000 \\
Maximum response length (RLCR) & 4,096 \\
Rollout temperature & 1.0 \\
Clip ratio (lower / upper) & 0.20 / 0.28 \\
\midrule
\multicolumn{2}{l}{\textit{Calibrator training}} \\
Representation layer & Final transformer block \\
Input dimension (8B / 14B) & 8,192 / 10,240 \\
MLP hidden-layer widths & 2,048 / 1,024 / 512 / 256 \\
Train / validation split & 90\% / 10\% by question \\
Loss & MSE \\
Optimizer & Adam \\
CoCal learning rate & $4 \times 10^{-4}$ \\
CoCal weight decay & 0  \\
CoCal training order & Chronological \\
Batch size & 256 \\
Checkpoint selection & Lowest binary validation MSE \\
Random seeds & 42, 137, 500, 700, 1024 \\
\midrule
\multicolumn{2}{l}{\textit{Evaluation }} \\
Maximum response length (AIME / AMC) & 16,384 \\
Maximum response length (Other datasets) & 4,096 \\
Sampling temperature & 0.7 \\
Top-$p$ & 0.8 \\
Top-$k$ & 20 \\
Presence penalty & 1.5 \\
\bottomrule
\end{tabular}
\end{table}

\section{Baseline Details}
\label{app:baseline_details}

\subsection{Training-Free Confidence Estimation}

\paragraph{Evaluation responses.}
\coco{}, SelfConsis, TokenProb, and P(True) are evaluated on the same fixed responses generated by the GRPO-trained policy. Verbal confidence is evaluated on responses generated in its confidence-reporting format. RLCR and DCPO modify the policy generation process during training, so they are evaluated on responses generated by their own trained policies.

\paragraph{SelfConsis.}
For an evaluated response $o$ and a pool of $M=10$ responses
$\{o^{(m)}\}_{m=1}^{M}$ sampled for the same question, confidence is computed as the fraction of responses consistent with the evaluated answer:
\begin{equation}
    c_{\mathrm{SC}}(o)
    =
    \frac{1}{M}
    \sum_{m=1}^{M}
    \mathbb{I}\!\left[
        a(o^{(m)}) \equiv a(o)
    \right],
\end{equation}
where $a(\cdot)$ denotes the extracted final answer and $\equiv$ denotes answer equivalence.
For mathematical tasks, equivalence is determined by deterministic answer matching; for factual QA, we use Qwen2.5-32B~\citep{qwen2025qwen25technicalreport} as a semantic-equivalence judge.

\paragraph{TokenProb.}
For a response $o=(o_1,\ldots,o_L)$, TokenProb uses the mean of its token-level generation probability as the confidence:
\begin{equation}
    c_{\mathrm{Tok}}(o)
    =
    \exp\left(
        \frac{1}{L}
        \sum_{m=1}^{L}
        \log
        \pi(o_m\mid q,o_{<m})
    \right).
\end{equation}
When present, the generated end-of-sequence token is included.
Probabilities are computed from the policy's raw logits rather than the temperature-adjusted sampling distribution.

\paragraph{P(True).}
Given question $q$ and its fixed response $o$, the model is prompted to judge whether the final answer is correct.
Let $p_{\mathrm{T}}$ and $p_{\mathrm{F}}$ denote the next-token probabilities assigned to \texttt{True} and \texttt{False}, respectively.
We compute
\begin{equation}
    c_{\text{P(True)}}(o)
    =
    \frac{p_{\mathrm{T}}}
    {p_{\mathrm{T}}+p_{\mathrm{F}}}.
\end{equation}

\subsection{Training-Based Confidence Estimation}

For RLCR and DCPO, we follow their original policy-optimization objectives and settings.
For all policy-optimization methods, we set the KL coefficient to zero following prior work~\citep{yu2026dapo,damani2026beyond,ma2026decoupling}.

\paragraph{RLCR.}
RLCR~\citep{damani2026beyond} generates a solution and answer, followed by uncertainty reasoning and a confidence score. For mathematical tasks, our implementation uses the following output format:
\begin{quote}
\ttfamily
Solution reasoning\\
\textbackslash boxed\{answer\}\\
<reason>Uncertainty reasoning</reason>\\
CONFIDENCE: score
\end{quote}
Given confidence $c_i\in[0,1]$ and verifier correctness $y_i\in\{0,1\}$, RLCR combines task correctness with a Brier-style calibration reward:
\begin{equation}
    \label{eq:rlcr_reward}
    R_i^{\mathrm{RLCR}} = y_i-(c_i-y_i)^2.
\end{equation}
The combined reward is normalized within each rollout group to obtain the GRPO advantage, which is applied to the entire generated sequence.

\paragraph{DCPO.}
DCPO~\citep{ma2026decoupling} removes the additional confidence-specific reasoning and separates task and calibration optimization across token subsets.
The task reward is response correctness,
\begin{equation}
    R_i^{r}=y_i.
\end{equation}
For confidence learning, DCPO constructs a target combining individual correctness with group accuracy:
\begin{equation}
    y_i^{\mathrm{DCPO}}
    =
    (1-\lambda)y_i
    +
    \lambda\bar{y},
    \qquad
    \bar{y}
    =
    \frac{1}{G}
    \sum_{j=1}^{G}y_j,
\end{equation}
and defines the calibration reward as
\begin{equation}
    R_i^{c}
    =
    -\left|
        c_i-y_i^{\mathrm{DCPO}}
    \right|.
\end{equation}
Here, $\lambda$ controls the contribution of group accuracy and is set to 0.5 as in its original implementation. Task and calibration rewards are normalized separately to obtain their respective advantages.
The task advantage is applied only to reasoning-and-answer tokens, while the calibration advantage is applied only to confidence tokens.
Thus, DCPO separates the direct token-level learning signals, but both objectives are still optimized through the shared policy parameters.

\paragraph{PostCal.}
PostCal uses the same architecture, representations, targets, and checkpoint criterion as CoCal, but trains on fresh final-policy responses with shuffled batches.
The main comparison uses the same 245,760-response raw budget as CoCal. After excluding truncated responses, PostCal retains 152,065 training and 16,968 validation responses. Table~\ref{tab:hyperparameters} lists its optimizer settings; Appendix~\ref{app:rq4_data_composition} details the budget comparison.

\section{Confidence Quality across Training Stages}
\label{app:training_stages}
\subsection{Earlier Training Stages}
Tables~\ref{tab:appendix_iter39} and~\ref{tab:appendix_iter79} report the 40- and 80-step results. Both \coco{} variants outperform RLCR and DCPO in ECE and AUROC at step 40. At step 80, RLCR achieves the lowest ECE, while \coco{} (binary) has the highest AUROC; however, RLCR's task accuracy remains substantially lower than that of \coco{}. \looseness=-1

\begin{table}[!htbp]
\centering
\caption{Accuracy and confidence quality on five mathematical datasets using Qwen3-8B after 40 training steps. Bold and underlined values indicate the best and second-best results, respectively.}
\label{tab:appendix_iter39}
\setlength{\tabcolsep}{2.6pt}
\renewcommand{\arraystretch}{1.08}
\resizebox{\textwidth}{!}{%
\begin{tabular}{l*{12}{c}}
\toprule
\multirow{2}{*}{Method} & \multicolumn{4}{c}{MATH500} & \multicolumn{4}{c}{AIME24} & \multicolumn{4}{c}{AIME25} \\
\cmidrule(lr){2-5}\cmidrule(lr){6-9}\cmidrule(lr){10-13}
& Acc.$\uparrow$ & MeanConf & ECE$\downarrow$ & AUROC$\uparrow$ & Acc.$\uparrow$ & MeanConf & ECE$\downarrow$ & AUROC$\uparrow$ & Acc.$\uparrow$ & MeanConf & ECE$\downarrow$ & AUROC$\uparrow$ \\
\midrule
RLCR & 85.8 & 0.963 & 0.098 & 0.747 & 37.0 & 0.762 & 0.392 & 0.860 & 25.3 & 0.784 & 0.530 & \underline{0.895} \\
DCPO & \underline{86.6} & 0.994 & 0.105 & 0.706 & \underline{37.7} & 0.937 & 0.556 & 0.766 & \underline{28.0} & 0.940 & 0.660 & 0.875 \\
\midrule
\coco{} (group) & \textbf{87.0} & 0.784 & \underline{0.093} & \underline{0.909} & \textbf{39.3} & 0.392 & \textbf{0.068} & \textbf{0.890} & \textbf{30.3} & 0.368 & \textbf{0.124} & 0.891 \\
\coco{} (binary) & \textbf{87.0} & 0.831 & \textbf{0.055} & \textbf{0.910} & \textbf{39.3} & 0.478 & \underline{0.126} & \underline{0.887} & \textbf{30.3} & 0.446 & \underline{0.166} & \textbf{0.902} \\
\bottomrule
\end{tabular}%
}

\vspace{2pt}

\resizebox{\textwidth}{!}{%
\begin{tabular}{l*{12}{c}}
\toprule
\multirow{2}{*}{Method} & \multicolumn{4}{c}{AMC23} & \multicolumn{4}{c}{AMC24} & \multicolumn{4}{c}{Macro Average} \\
\cmidrule(lr){2-5}\cmidrule(lr){6-9}\cmidrule(lr){10-13}
& Acc.$\uparrow$ & MeanConf & ECE$\downarrow$ & AUROC$\uparrow$ & Acc.$\uparrow$ & MeanConf & ECE$\downarrow$ & AUROC$\uparrow$ & Acc.$\uparrow$ & MeanConf & ECE$\downarrow$ & AUROC$\uparrow$ \\
\midrule
RLCR & 64.8 & 0.918 & 0.271 & 0.716 & 56.7 & 0.853 & 0.286 & 0.808 & 53.9 & 0.856 & 0.315 & 0.805 \\
DCPO & \underline{78.0} & 0.980 & 0.198 & 0.745 & \underline{58.4} & 0.955 & 0.362 & 0.718 & \underline{57.7} & 0.961 & 0.376 & 0.762 \\
\midrule
\coco{} (group) & \textbf{78.8} & 0.639 & \underline{0.150} & \underline{0.847} & \textbf{64.2} & 0.564 & \underline{0.105} & \underline{0.841} & \textbf{59.9} & 0.549 & \underline{0.108} & \underline{0.876} \\
\coco{} (binary) & \textbf{78.8} & 0.708 & \textbf{0.081} & \textbf{0.855} & \textbf{64.2} & 0.652 & \textbf{0.056} & \textbf{0.870} & \textbf{59.9} & 0.623 & \textbf{0.097} & \textbf{0.885} \\
\bottomrule
\end{tabular}%
}

\end{table}

\begin{table}[!htbp]
\centering
\caption{Accuracy and confidence quality on five mathematical datasets using Qwen3-8B after 80 training steps. Bold and underlined values indicate the best and second-best results, respectively.}
\label{tab:appendix_iter79}
\setlength{\tabcolsep}{2.6pt}
\renewcommand{\arraystretch}{1.08}
\resizebox{\textwidth}{!}{%
\begin{tabular}{l*{12}{c}}
\toprule
\multirow{2}{*}{Method} & \multicolumn{4}{c}{MATH500} & \multicolumn{4}{c}{AIME24} & \multicolumn{4}{c}{AIME25} \\
\cmidrule(lr){2-5}\cmidrule(lr){6-9}\cmidrule(lr){10-13}
& Acc.$\uparrow$ & MeanConf & ECE$\downarrow$ & AUROC$\uparrow$ & Acc.$\uparrow$ & MeanConf & ECE$\downarrow$ & AUROC$\uparrow$ & Acc.$\uparrow$ & MeanConf & ECE$\downarrow$ & AUROC$\uparrow$ \\
\midrule
RLCR & \underline{88.0} & 0.897 & \underline{0.058} & 0.835 & 37.7 & 0.356 & \textbf{0.092} & \textbf{0.936} & 28.3 & 0.330 & \textbf{0.049} & \textbf{0.964} \\
DCPO & 87.0 & 0.977 & 0.069 & 0.684 & \underline{41.7} & 0.754 & 0.338 & 0.870 & \underline{33.7} & 0.741 & 0.405 & 0.930 \\
\midrule
\coco{} (group) & \textbf{89.0} & 0.822 & 0.075 & \underline{0.898} & \textbf{44.0} & 0.414 & \underline{0.104} & \underline{0.875} & \textbf{36.7} & 0.389 & \underline{0.099} & 0.929 \\
\coco{} (binary) & \textbf{89.0} & 0.860 & \textbf{0.051} & \textbf{0.905} & \textbf{44.0} & 0.547 & 0.160 & 0.855 & \textbf{36.7} & 0.517 & 0.185 & \underline{0.943} \\
\bottomrule
\end{tabular}%
}

\vspace{2pt}

\resizebox{\textwidth}{!}{%
\begin{tabular}{l*{12}{c}}
\toprule
\multirow{2}{*}{Method} & \multicolumn{4}{c}{AMC23} & \multicolumn{4}{c}{AMC24} & \multicolumn{4}{c}{Macro Average} \\
\cmidrule(lr){2-5}\cmidrule(lr){6-9}\cmidrule(lr){10-13}
& Acc.$\uparrow$ & MeanConf & ECE$\downarrow$ & AUROC$\uparrow$ & Acc.$\uparrow$ & MeanConf & ECE$\downarrow$ & AUROC$\uparrow$ & Acc.$\uparrow$ & MeanConf & ECE$\downarrow$ & AUROC$\uparrow$ \\
\midrule
RLCR & 68.3 & 0.760 & 0.187 & 0.751 & 58.4 & 0.639 & 0.099 & \textbf{0.880} & 56.1 & 0.597 & \textbf{0.097} & \underline{0.873} \\
DCPO & \underline{83.8} & 0.916 & \underline{0.096} & 0.748 & \underline{64.9} & 0.842 & 0.193 & 0.766 & \underline{62.2} & 0.846 & 0.220 & 0.799 \\
\midrule
\coco{} (group) & \textbf{84.5} & 0.677 & 0.168 & \underline{0.807} & \textbf{65.6} & 0.607 & \textbf{0.075} & 0.801 & \textbf{63.9} & 0.582 & \underline{0.104} & 0.862 \\
\coco{} (binary) & \textbf{84.5} & 0.771 & \textbf{0.082} & \textbf{0.835} & \textbf{65.6} & 0.710 & \underline{0.082} & \underline{0.840} & \textbf{63.9} & 0.681 & 0.112 & \textbf{0.876} \\
\bottomrule
\end{tabular}%
}
\end{table}

\subsection{Learning from Early Rollout Data}
\begin{figure}
    \centering
    \includegraphics[width=\linewidth]{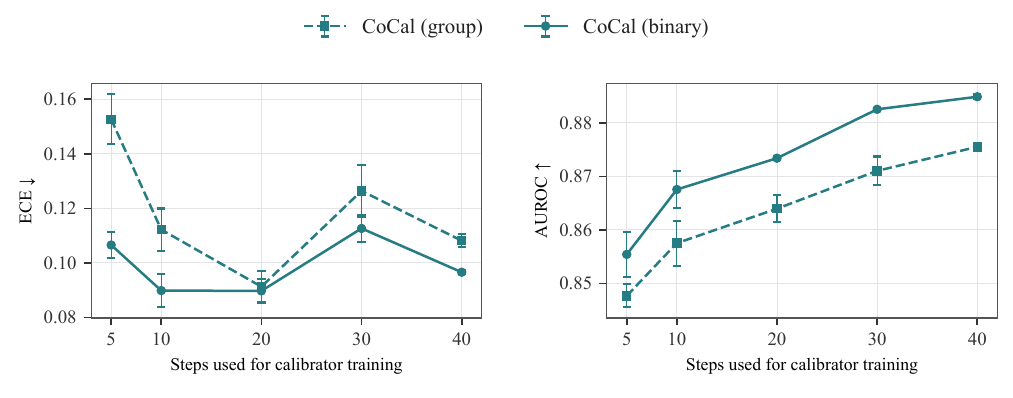}
    \caption{Confidence estimation using training data produced by early steps, evaluated on the fixed step-40 GRPO policy. Points average five seeds and five mathematical datasets; error bars show seed standard deviations.}
    \label{fig:early_step_cocal}
\end{figure}

Figure~\ref{fig:training_dynamics} shows that RLCR and DCPO still have relatively weak confidence estimates at step 40 and improve substantially by step 80. Since policy checkpoints were saved only every 40 steps due to storage constraints, we further probe earlier learning by training \coco{} on rollout prefixes from the first 5, 10, 20, 30, or 40 steps. For each prefix, both training and validation are restricted to the corresponding rollout data while preserving the question-level split, and all companions are evaluated on the same step-40 policy responses. As shown in Figure~\ref{fig:early_step_cocal}, \coco{} (binary) already achieves an ECE of 0.090 and an AUROC of 0.868 with only the first 10 steps of rollout data, indicating that strong confidence estimation can emerge from very limited early training experience.

\subsection{Full Reliability Diagrams}
\begin{figure}[p]
\centering
\includegraphics[width=\linewidth]{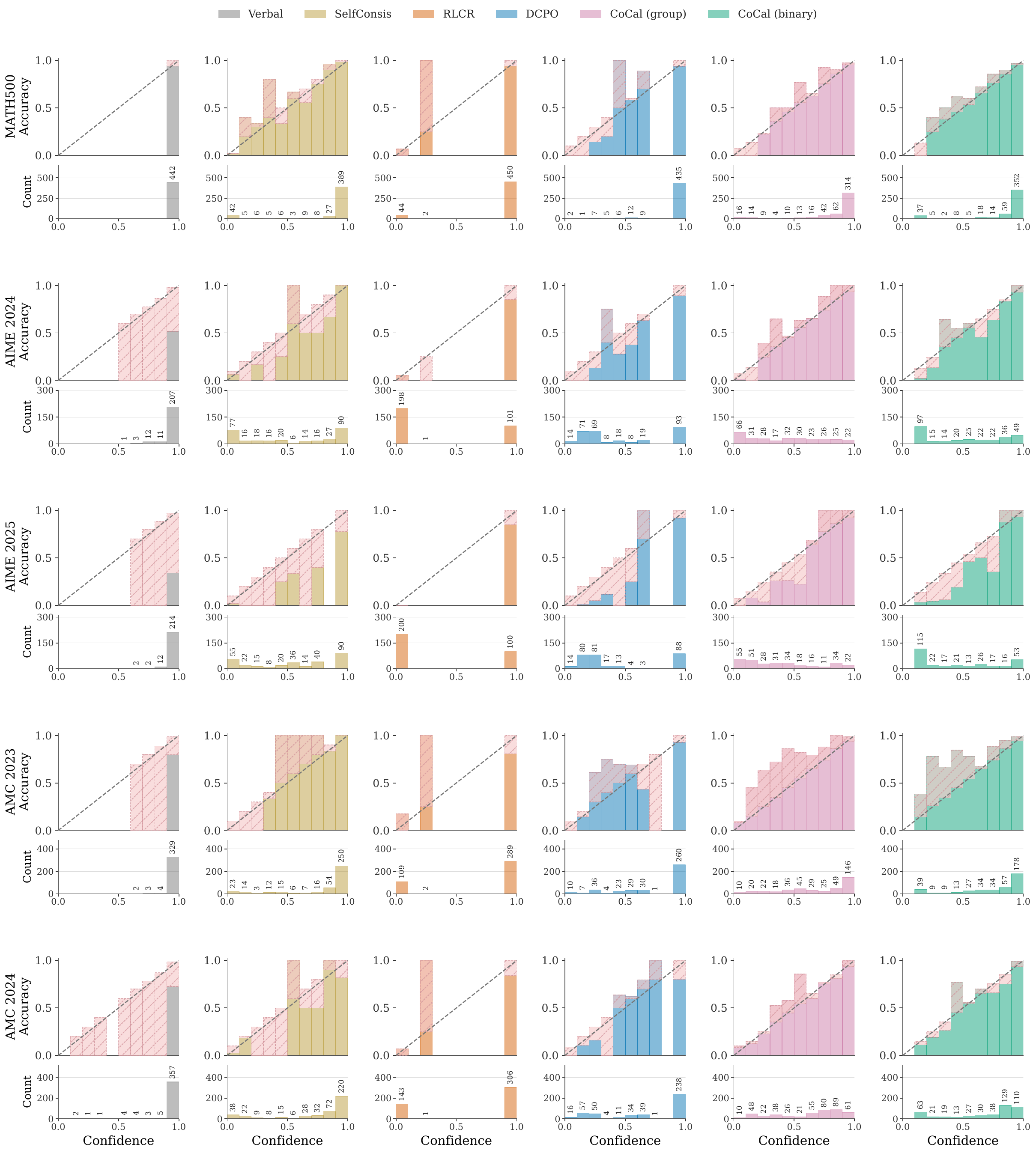}
\caption{Reliability diagrams for Qwen3-8B after 120 policy-training steps on all five mathematical datasets. Colored bars show accuracy within nonempty confidence bins; hatched regions show the absolute gap to the actual bin-mean confidence. The dashed diagonal is an ideal-calibration reference. Lower panels report bin counts.}
\label{fig:rq2_reliability_all}
\end{figure}

Figure~\ref{fig:rq2_reliability_all} extends the main-text analysis to all five mathematical datasets and shows similar patterns: Verbal and DCPO concentrate confidence toward the high end, often exceeding empirical accuracy, while RLCR favors near-zero or near-one estimates. Both \coco{} variants distribute confidence across more bins, providing finer-grained estimates, although calibration varies across datasets. CoCal uses seed 42 to display integer bin counts; empty bins are left blank.

\section{Cross-Iteration and Cross-Policy Transfer}
In Figure~\ref{fig:rq3_cross_iter}, source $X$ denotes the rollout steps used to train the companion, while target $Y$ denotes the policy checkpoint on which it is evaluated. Off-diagonal cells therefore measure cross-iteration transfer without retraining, yielding ECE of 0.092--0.119 and AUROC of 0.869--0.893.

For cross-policy transfer (Figure~\ref{fig:rq3_policy_transfer}), we apply GRPO-trained companions to matched-stage DCPO and RLCR policies with the same backbone. For DCPO, we retain the response endpoint when extracting features. For RLCR, we extract features only from the task response, stopping before the first \texttt{<reason>} tag and excluding uncertainty reasoning and confidence text.

\section{Comparison with PostCal}
\label{app:comparision_with_postcal}
\subsection{Matched Budgets and Training-data Composition}
\label{app:rq4_data_composition}

\begin{figure}[!htbp]
\centering
\includegraphics[width=0.85\linewidth]{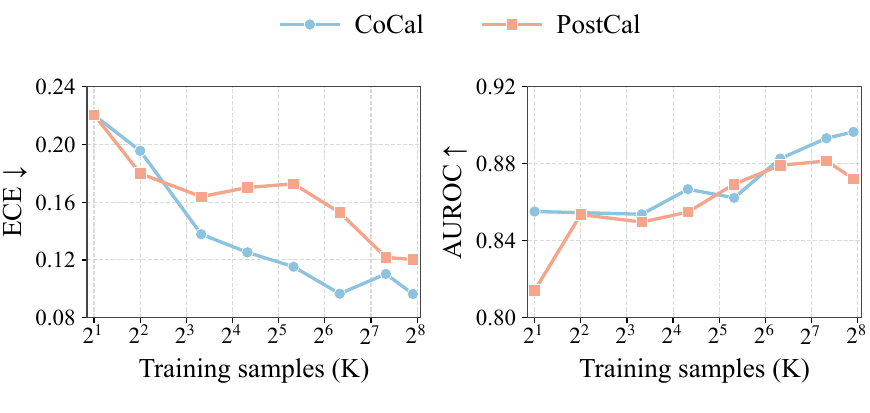}
\caption{Confidence performance across matched rollout-data budgets. Both methods use binary supervision; results average five mathematical datasets.}
\label{fig:rq4_cocal_vs_posthoc}
\end{figure}

We compare \coco{} and PostCal under binary supervision across matched rollout-data budgets corresponding to the first 1/2/5/10/20/40/80/120 RLVR steps. For each budget, \coco{} uses the rollouts collected during RLVR, while PostCal resamples responses from the final step-120 policy using the same questions and the same number of responses per question. Both methods discard truncated responses and are evaluated on the same step-120 policy responses. This study matches raw sampling budgets, as in the main comparison. Its full-budget binary results use the same checkpoints as the main tables. Each budget uses a fixed, question-disjoint validation pool.

\begin{figure}[!htbp]
\centering
\includegraphics[width=\linewidth]{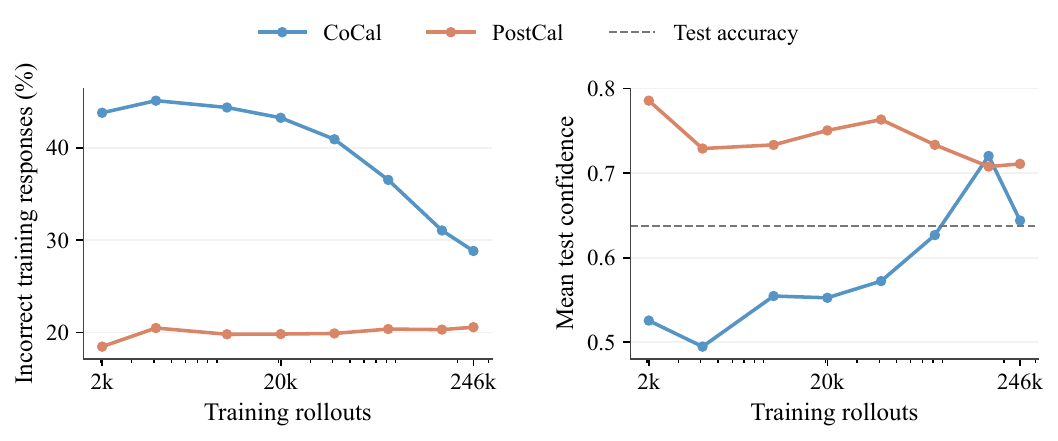}
\caption{Training-data composition and mean test confidence under matched raw rollout budgets. Both methods use binary supervision. Left: the fraction of incorrect effective training responses, after excluding validation and truncated responses. Right: mean confidence on the fixed 120-step test responses, averaged over five companion seeds and five mathematical datasets. The dashed line marks macro task accuracy.}
\label{fig:rq4_error_confidence}
\end{figure}

As shown in Figure~\ref{fig:rq4_cocal_vs_posthoc}, \coco{} achieves lower ECE and higher AUROC in seven of eight budgets. Its training data also contain a higher fraction of incorrect responses across budgets and generally lead to lower test confidence (Figure~\ref{fig:rq4_error_confidence}). This suggests that exposure to errors from earlier policy stages may help reduce overconfidence, although error frequency is only one of several differences between the two training-data distributions.

\subsection{Computational-cost Accounting}
\label{app:cost}
\label{app:rq4_timing}
PostCal sampling time is measured on 30,720 question occurrences from DeepScaleR using vllm on one B200 GPU, with eight responses per occurrence. Both \coco{} and PostCal use the same raw rollout budget of 245,760 responses, corresponding to 30,720 question occurrences with eight responses sampled per question. We measure LLM generation time on all 500 MATH500 questions using vLLM 0.9.2 on a single NVIDIA B200 GPU with BF16 precision and tensor parallelism of 1. All questions are submitted in one batch, with at most 64 concurrent sequences, CUDA graphs enabled, and prefix caching disabled. We report total generation wall-clock time divided by 500, using one response per question for CoCal and PostCal and ten for SelfConsis. Timing excludes model loading, compilation, warm-up, tokenization, and answer scoring, including SelfConsis answer-equivalence computation. MLP latency is averaged over 1,000 batch-size-one predictions.

\section{Further Analysis}
\label{app:further_analysis}
\subsection{Model Scale}
\label{app:model_scale}
Table~\ref{tab:main_iter119_14b_compact} reports Qwen3-14B results on mathematical datasets. Both \coco{} variants preserve task accuracy and outperform RLCR and DCPO in ECE and AUROC; \coco{} (binary) performs best on both metrics.

\begin{table}[!htbp]
\centering
\caption{Accuracy and confidence quality on five mathematical datasets using Qwen3-14B after 120 training steps. Base + Verbal uses the pre-RL model; other training-free baselines use the GRPO-trained model. Bold and underlined values indicate the best and second-best results, respectively.}
\label{tab:main_iter119_14b_compact}
\setlength{\tabcolsep}{2.6pt}
\renewcommand{\arraystretch}{1.08}
\resizebox{\textwidth}{!}{%
\begin{tabular}{l*{12}{c}}
\toprule
\multirow{2}{*}{Method}
& \multicolumn{4}{c}{MATH500}
& \multicolumn{4}{c}{AIME24}
& \multicolumn{4}{c}{AIME25} \\
\cmidrule(lr){2-5}\cmidrule(lr){6-9}\cmidrule(lr){10-13}
& Acc.$\uparrow$ & MeanConf & ECE$\downarrow$ & AUROC$\uparrow$
& Acc.$\uparrow$ & MeanConf & ECE$\downarrow$ & AUROC$\uparrow$
& Acc.$\uparrow$ & MeanConf & ECE$\downarrow$ & AUROC$\uparrow$ \\
\midrule
Base + Verbal & 84.0 & 0.993 & 0.151 & 0.717 & 30.7 & 0.964 & 0.654 & 0.728 & 23.7 & 0.967 & 0.729 & 0.800 \\
\midrule
Verbal & 88.8 & 0.991 & 0.103 & 0.747 & 44.3 & 0.949 & 0.506 & 0.712 & 33.7 & 0.961 & 0.624 & 0.820 \\
SelfConsis & \textbf{90.8} & 0.925 & \textbf{0.045} & 0.863 & \textbf{49.0} & 0.565 & 0.096 & \textbf{0.939} & \textbf{37.0} & 0.548 & 0.178 & 0.868 \\
TokenProb & \textbf{90.8} & 0.958 & \underline{0.050} & 0.813 & \textbf{49.0} & 0.886 & 0.396 & 0.847 & \textbf{37.0} & 0.867 & 0.497 & 0.897 \\
P(True) & \textbf{90.8} & 0.925 & 0.078 & 0.842 & \textbf{49.0} & 0.667 & 0.222 & 0.844 & \textbf{37.0} & 0.601 & 0.280 & 0.869 \\
\midrule
RLCR & \underline{90.2} & 0.938 & 0.060 & 0.754 & \underline{47.0} & 0.507 & 0.094 & \underline{0.921} & 35.0 & 0.407 & \underline{0.068} & \textbf{0.947} \\
DCPO & 89.8 & 0.933 & 0.079 & 0.727 & 45.0 & 0.512 & 0.133 & 0.869 & \underline{36.0} & 0.419 & 0.115 & 0.882 \\
\midrule
\coco{} (group) & \textbf{90.8} & 0.842 & 0.072 & \underline{0.904} & \textbf{49.0} & 0.441 & \textbf{0.071} & 0.908 & \textbf{37.0} & 0.389 & \textbf{0.063} & 0.903 \\
\coco{} (binary) & \textbf{90.8} & 0.866 & \underline{0.050} & \textbf{0.908} & \textbf{49.0} & 0.550 & \underline{0.074} & 0.907 & \textbf{37.0} & 0.492 & 0.145 & \underline{0.914} \\
\bottomrule
\end{tabular}%
}

\resizebox{\textwidth}{!}{%
\begin{tabular}{l*{12}{c}}
\toprule
\multirow{2}{*}{Method}
& \multicolumn{4}{c}{AMC23}
& \multicolumn{4}{c}{AMC24}
& \multicolumn{4}{c}{Macro Average} \\
\cmidrule(lr){2-5}\cmidrule(lr){6-9}\cmidrule(lr){10-13}
& Acc.$\uparrow$ & MeanConf & ECE$\downarrow$ & AUROC$\uparrow$
& Acc.$\uparrow$ & MeanConf & ECE$\downarrow$ & AUROC$\uparrow$
& Acc.$\uparrow$ & MeanConf & ECE$\downarrow$ & AUROC$\uparrow$ \\
\midrule
Base + Verbal & 73.8 & 0.986 & 0.247 & 0.704 & 50.9 & 0.978 & 0.464 & 0.712 & 52.6 & 0.978 & 0.449 & 0.732 \\
\midrule
Verbal & 85.0 & 0.981 & 0.131 & 0.802 & 63.8 & 0.964 & 0.326 & 0.749 & 63.1 & 0.969 & 0.338 & 0.766 \\
SelfConsis & \textbf{86.5} & 0.842 & \textbf{0.060} & \textbf{0.994} & \textbf{67.8} & 0.796 & 0.152 & 0.778 & \textbf{66.2} & 0.735 & 0.106 & 0.888 \\
TokenProb & \textbf{86.5} & 0.935 & \underline{0.070} & 0.856 & \textbf{67.8} & 0.924 & 0.246 & 0.748 & \textbf{66.2} & 0.914 & 0.252 & 0.832 \\
P(True) & \textbf{86.5} & 0.895 & 0.140 & 0.795 & \textbf{67.8} & 0.838 & 0.214 & 0.841 & \textbf{66.2} & 0.785 & 0.187 & 0.838 \\
\midrule
RLCR & \underline{86.2} & 0.860 & 0.071 & 0.867 & 63.6 & 0.781 & 0.158 & 0.794 & \underline{64.4} & 0.699 & 0.090 & 0.857 \\
DCPO & 82.5 & 0.840 & 0.158 & 0.714 & \underline{65.8} & 0.784 & 0.154 & 0.773 & 63.8 & 0.697 & 0.128 & 0.793 \\
\midrule
\coco{} (group) & \textbf{86.5} & 0.737 & 0.128 & 0.874 & \textbf{67.8} & 0.613 & \underline{0.104} & \underline{0.857} & \textbf{66.2} & 0.604 & \underline{0.088} & \underline{0.889} \\
\coco{} (binary) & \textbf{86.5} & 0.794 & 0.071 & \underline{0.882} & \textbf{67.8} & 0.693 & \textbf{0.071} & \textbf{0.874} & \textbf{66.2} & 0.679 & \textbf{0.082} & \textbf{0.897} \\
\bottomrule
\end{tabular}%
}
\end{table}

\subsection{Token Aggregation}
\label{app:token_aggregation}
\label{app:response_representations}
Table~\ref{tab:response_features} compares different token aggregation strategies under binary supervision. Overall, \coco{} is robust to the choice of response representation, with all three variants achieving strong confidence estimation. The final-token state performs consistently well, achieving the best ECE and AUROC on Qwen3-8B and the best AUROC on Qwen3-14B. Mean pooling alone yields weaker discrimination on both backbones, while concatenating the two remains competitive and achieves the lowest ECE on Qwen3-14B. These results suggest that the final-token state already captures strong confidence signals, with response-level pooling providing complementary information in some settings.

\begin{table}[!htbp]
\centering
\captionsetup{font=small,skip=4pt}
\caption{Confidence performance with different token aggregations, averaged over five seeds and five mathematical datasets.}
\label{tab:response_features}
\small
\setlength{\tabcolsep}{4pt}
\begin{tabular}{lcccc}
\toprule
\multirow{2}{*}{Feature} & \multicolumn{2}{c}{Qwen3-8B} & \multicolumn{2}{c}{Qwen3-14B} \\
\cmidrule(lr){2-3}\cmidrule(lr){4-5}
& ECE$\downarrow$ & AUROC$\uparrow$ & ECE$\downarrow$ & AUROC$\uparrow$ \\
\midrule
Last token & \textbf{0.093} & \textbf{0.900} & 0.087 & \textbf{0.898} \\
Response mean & 0.095 & 0.871 & 0.104 & 0.869 \\
Last + mean & 0.096 & 0.896 & \textbf{0.082} & 0.897 \\
\bottomrule
\end{tabular}
\end{table}

\subsection{Layer Selection}
\label{app:layer_selection}
\begin{figure}[!htbp]
\centering
\includegraphics[width=0.85\textwidth]{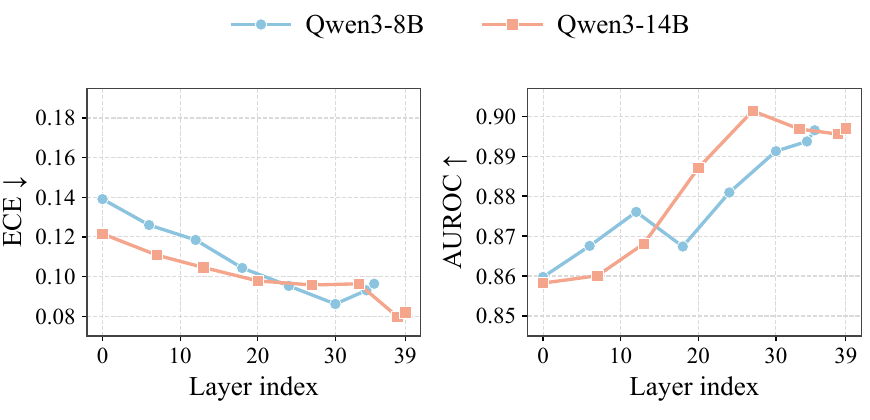}
\caption{Average confidence estimation performance using internal states from different layers on mathematical datasets.}
\label{fig:layer_depth}
\end{figure}

Each companion uses binary supervision and the concatenated representation in Equation~\ref{eq:response_features}; only the extraction layer varies. Layer indices are zero-based, ending at 35 for Qwen3-8B and 39 for Qwen3-14B. As shown in Figure~\ref{fig:layer_depth}, middle-to-late layers consistently provide strong confidence signals on both backbones, while early layers perform noticeably worse. The lowest ECE occurs at layers 30 and 38 for Qwen3-8B and Qwen3-14B, respectively, while AUROC peaks at layers 35 and 27. Performance varies only modestly across later layers, suggesting that confidence-relevant information is broadly available in deeper representations rather than concentrated at a single layer. We therefore use the final layer as a simple and robust default.

\Needspace{20\baselineskip}
\subsection{Pre-generation Confidence Estimation}
\label{app:pre_generation}

We examine whether \coco{} can estimate confidence before response generation. The predictor takes the final prompt-token state as input. Since no individual response is available at this stage, we use the mean correctness of the original $K=8$ rollout group as the supervision target, corresponding to question-level reliability. Unlike post-generation group supervision, this target is not multiplied by individual response correctness.

To approximate historical prompt states, we use the policies trained for 40, 80, and 120 steps for rollouts from steps 1--40, 41--80, and 81--120, respectively. Training follows the same chronological protocol as \coco{}, with no shuffling and question-level train/validation splits. We use the group variant's optimizer settings in Table~\ref{tab:hyperparameters}, and select checkpoints by validation MSE against group accuracy.

Table~\ref{tab:pre_generation} compares pre-generation confidence with post-generation \coco{} (group) on the same evaluation responses. Prompt-only states provide informative confidence estimates, but post-generation confidence achieves substantially lower ECE and higher AUROC on both domains. This gap is expected because pre-generation confidence lacks response-specific information: it has no access to the generated reasoning or answer, and all responses to the same question receive the same confidence score, limiting discrimination between correct and incorrect generations. Post-generation \coco{}, in contrast, conditions confidence on each generated response and can capture response-specific variation. Since the two settings also differ in supervision and historical-state extraction, this comparison does not isolate the effect of response information alone.
\begin{table}[t]
\centering
\caption{Pre- and post-generation confidence performance of \coco{} on Qwen3-8B after 120 policy-training steps. Results average five seeds and five datasets.}
\label{tab:pre_generation}
\setlength{\tabcolsep}{6pt}
\begin{tabular}{lcccc}
\toprule
\multirow{2}{*}{Setting}
& \multicolumn{2}{c}{Math}
& \multicolumn{2}{c}{HonestyBench} \\
\cmidrule(lr){2-3}\cmidrule(lr){4-5}
& ECE$\downarrow$ & AUROC$\uparrow$
& ECE$\downarrow$ & AUROC$\uparrow$ \\
\midrule
Pre-generation          & 0.124 & 0.785 & 0.268 & 0.650 \\
Post-generation (group) & 0.103 & 0.884 & 0.104 & 0.725 \\
\bottomrule
\end{tabular}
\end{table}

\begin{table}[t]
\centering
\caption{Mean time to confidence on eight questions per dataset.}
\label{tab:pre_generation_latency}
\setlength{\tabcolsep}{7pt}
\begin{tabular}{lrrr}
\toprule
Dataset & Pre (ms) & Post (s) & Post/Pre \\
\midrule
MATH500   & 38.78 & 58.46  & $1508\times$ \\
AIME 2024 & 43.76 & 230.11 & $5258\times$ \\
AIME 2025 & 41.78 & 184.56 & $4417\times$ \\
AMC 2023  & 24.71 & 62.94  & $2548\times$ \\
AMC 2024  & 25.22 & 79.72  & $3162\times$ \\
\bottomrule
\end{tabular}
\end{table}

\paragraph{Time to confidence.}
To fairly compare pre-generation and post-generation confidence, we use the same Transformers inference pipeline rather than using vllm for both settings, allowing the pre-generation route to stop after prompt encoding without decoding a response. We measure eight randomly sampled questions from each mathematical dataset using the same Qwen3-8B policy on a single B200 GPU with Transformers 4.53.1, BF16, SDPA, and batch size one. Pre-generation latency includes prompt encoding and MLP prediction, whereas post-generation latency includes cached autoregressive decoding, final-state processing, and MLP prediction. Model loading, tokenization, and warm-up are excluded.

Table~\ref{tab:pre_generation_latency} shows that pre-generation confidence is available within 25--44\,ms, while post-generation confidence requires tens to hundreds of seconds depending on response length. This reduces time to confidence by roughly three to four orders of magnitude, enabling confidence-aware decisions such as abstention, retrieval, or allocating additional computation before response generation begins.

\end{document}